# Speeding Up Latent Variable Gaussian Graphical Model Estimation via Nonconvex Optimizations

Pan Xu[*] and Jian Ma[†] and Quanquan Gu[‡]

February 24, 2017

**Abstract**

We study the estimation of the latent variable Gaussian graphical model (LVGGM), where the precision matrix is the superposition of a sparse matrix and a low-rank matrix. In order to speed up the estimation of the sparse plus low-rank components, we propose a sparsity constrained maximum likelihood estimator based on matrix factorization, and an efficient alternating gradient descent algorithm with hard thresholding to solve it. Our algorithm is orders of magnitude faster than the convex relaxation based methods for LVGGM. In addition, we prove that our algorithm is guaranteed to linearly converge to the unknown sparse and low-rank components up to the optimal statistical precision. Experiments on both synthetic and genomic data demonstrate the superiority of our algorithm over the state-of-the-art algorithms and corroborate our theory.

## 1 Introduction

For a $d$-dimensional Gaussian graphical model (i.e., multivariate Gaussian distribution) $N(\mathbf{0}, \mathbf{\Sigma}^*)$, the inverse of covariance matrix $\mathbf{\Omega}^* = (\mathbf{\Sigma}^*)^{-1}$ (also known as the precision matrix or concentration matrix) measures the conditional dependence relationship between marginal random variables (Lauritzen, 1996). When the number of observations is comparable to the ambient dimension of the Gaussian graphical model, additional structural assumptions are needed for consistent estimation. Sparsity is one of the most common structures imposed on the precision matrix in Gaussian graphical models (GGM), because it gives rise to a graph, which characterizes the conditional dependence of the marginal variables. The problem of estimating the sparse precision matrix in Gaussian graphical models has been studied by a large body of literature (Meinshausen and Bühlmann, 2006; Rothman et al., 2008; Friedman et al., 2008; Ravikumar et al., 2011; Cai et al., 2011). However, the real world data may not follow a sparse GGM, especially when some of the variables are unobservable.

---
[*]Department of Systems and Information Engineering, University of Virginia, Charlottesville, VA 22904, USA; e-mail: `px3ds@virginia.edu`

[†]Computational Biology Department, School of Computer Science, Carnegie Mellon University, Pittsburgh, PA 15213, USA; e-mail: `jianma@cs.cmu.edu`

[‡]Department of Systems and Information Engineering, Department of Computer Science, University of Virginia, Charlottesville, VA 22904, USA; e-mail: `qg5w@virginia.edu`

To alleviate this problem, the latent variable Gaussian graphical model (LVGGM) (Chandrasekaran et al., 2010; Meng et al., 2014) has been proposed, where the precision matrix of the observed variables is conditionally sparse given the latent variables (i.e., unobserved), but marginally not sparse. It is well-known that in LVGGM, the precision matrix $\mathbf{\Omega}^*$ can be represented as the superposition of a sparse matrix $\mathbf{S}^*$ and a low-rank matrix $\mathbf{L}^*$, where the latent variables contribute to the low rank component in the precision matrix. In other words, we have $\mathbf{\Omega}^* = \mathbf{S}^* + \mathbf{L}^*$.

In the learning problem of LVGGM, the goal is to estimate both the unknown sparse component $\mathbf{S}^*$ and the low-rank component $\mathbf{L}^*$ of the precision matrix simultaneously. In the seminal work, Chandrasekaran et al. (2010) proposed a maximum-likelihood estimator based on $\ell_1$ norm penalty on the sparse matrix and nuclear norm penalty on the low-rank matrix, and proved the model selection consistency for LVGGM estimation. Meng et al. (2014) studied a similar penalized estimator, and derived Frobenius norm error bounds based on the restricted strong convexity (Negahban et al., 2009) and the structural Fisher incoherence condition between the sparse and low-rank components. Both of these two methods for LVGGM estimation are based on a penalized convex optimization problem. Due to the nuclear norm penalty, they need to do singular value decomposition (SVD) to solve the proximal mapping of nuclear norm at each iteration, which results in an extremely high time complexity of $O(d^3)$. When $d$ is large as often in the high dimensional setting, the convex relaxation based methods are computationally intractable.

In this paper, we aim to speed up learning LVGGM without paying the computational price caused by the penalties (especially the nuclear norm penalty). To avoid the penalty, we propose a novel sparsity constrained maximum likelihood estimator for LVGGM based on matrix factorization. More specifically, inspired by the recent work on matrix completion (Jain et al., 2013; Hardt, 2014; Zhao et al., 2015; Zheng and Lafferty, 2015; Chen and Wainwright, 2015; Tu et al., 2015), to avoid the singular value decomposition at each iteration, we propose to reparameterize the low-rank component $\mathbf{L}$ in the precision matrix as the product of smaller matrices, i.e., $\mathbf{L} = \mathbf{Z}\mathbf{Z}^\top$, where $\mathbf{Z} \in \mathbb{R}^{d \times r}$ and $r \leq d$ is the number of latent variables. This factorization of $\mathbf{L}$ captures the intrinsic low-rank structure of $\mathbf{L}$, and automatically ensures the low-rankness of $\mathbf{L}$. We propose an estimator based on minimizing resulting nonconvex negative log-likelihood function under sparsity constraint, and an alternating gradient descent with hard thresholding to solve it. Our algorithm significantly reduces the per-iteration time complexity from $O(d^3)$ to $O(d^2 r)$, which greatly reduces the computation cost and scales up the estimation problem for LVGGM.

We prove that, provided that the initial points are sufficiently close to the sparse and low-rank components of the unknown precision matrix, the output of our algorithm is guaranteed to linearly converge to the unknown parameters up to the optimal statistical error. In particular, the estimators from our algorithm for LVGGM attain $\max\{O_p(\sqrt{s^* \log d/n}), O_p(\sqrt{rd/n})\}$ statistical rate of convergence in terms of Frobenius norm, where $s^*$ is the conditional sparsity of the precision matrix (i.e., sparsity of $\mathbf{S}^*$), and $r$ is the number of latent variables (i.e., rank of $\mathbf{L}^*$). This matches the minimax optimal convergence rate for LVGGM in Chandrasekaran et al. (2010); Agarwal et al. (2012a); Meng et al. (2014). To ensure the initial points satisfy the aforementioned closeness condition, we also present an initialization algorithm, which guarantees that the generated initial points meet the requirement.

It is also worth noting that, although our estimator and algorithm is designed for LVGGM, it is directly applicable to the Gaussian graphical model where the precision matrix is the sum of a sparse matrix and a low-rank matrix. And the theoretical guarantees of our algorithm still hold. Thorough experiments on both synthetic and real-world breast cancer datasets verify the effectiveness of our algorithm.

The remainder of this paper is organized as follows: In Section 2, we briefly review existing work that is relevant to our study. We present our estimator and algorithm in detail in Section 3, and the main theory in Section 4. In Section 5, we compare the proposed algorithm with the state-of-the-art algorithms on both synthetic data and real-world breast cancer data. Finally, we conclude this paper in Section 6.

**Notation** For a pair of matrices $\mathbf{A}, \mathbf{B}$ with commensurate dimensions, we let $\langle \mathbf{A}, \mathbf{B} \rangle = \text{tr}(\mathbf{A}^\top \mathbf{B})$ denote the inner product and let $\mathbf{A} \otimes \mathbf{B}$ denote the Kronecker product between them. For a matrix $\mathbf{A} \in \mathbb{R}^{d \times d}$, we denote its (ordered) singular values by $\sigma_1(\mathbf{A}) \geq \sigma_2(\mathbf{A}) \geq \ldots \geq \sigma_d(\mathbf{A}) \geq 0$. For a square matrix $\mathbf{A}$, we denote by $\mathbf{A}^{-1}$ the inverse of $\mathbf{A}$, and denote by $|\mathbf{A}|$ its determinant. We use the notation $\|\cdot\|$ for various types of matrix norms, including the spectral norm $\|\mathbf{A}\|_2 = \max_j \sigma_j(\mathbf{A})$, and the Frobenius norm $\|\mathbf{A}\|_F = \sqrt{\text{tr}(\mathbf{A}^\top \mathbf{A})} = \sqrt{\sum_{j=1}^m \sigma_j(\mathbf{A})^2}$. Also we have $\|\mathbf{A}\|_{0,0} = \sum_{i,j} \mathbb{1}(A_{ij} \neq 0)$, $\|\mathbf{A}\|_{\infty,\infty} = \max_{1 \leq i,j \leq m} |A_{ij}|$. $\|\mathbf{A}\|_{1,1} = \sum_{i,j=1}^m |A_{ij}|$.

## 2 Related Work

Precision matrix estimation in sparse Gaussian graphical models (GGM) is commonly formulated as a penalized maximum likelihood estimation problem with $\ell_1$ norm regularization (Friedman et al., 2008; Rothman et al., 2008; Ravikumar et al., 2011), which is also known as graphical Lasso. Due to the complex dependency among marginal variables in many applications, sparsity assumption on the precision matrix often does not hold. To relax this assumption, Yin and Li (2011); Cai et al. (2012) proposed the conditional Gaussian graphical model (cGGM) and Yuan and Zhang (2014) proposed the partial Gaussian graphical model (pGGM), both of which impose blockwise sparsity on the precision matrix and estimate multiple blocks therein. Despite a good interpretation of these models, they need to access both the observed variables as well as the latent variables for estimation. Another alternative is the latent variable Gaussian graphical model (LVGGM), which was proposed in Chandrasekaran et al. (2010), and later investigated in Agarwal et al. (2012a); Meng et al. (2014). Compared with cGGM and pGGM, the estimation of LVGGM does not need to access the latent variables and therefore is more flexible.

Existing algorithms for estimating LVGGM are based on convex relaxation methods using $\ell_1$ norm penalty and nuclear norm penalty. For instance, Chandrasekaran et al. (2010) proposed to use log-determinant proximal point algorithm (Wang et al., 2010) for LVGGM estimation. Ma et al. (2013); Meng et al. (2014) proposed to use alternating direction methods of multipliers (ADMM) to accelerate the estimation of LVGGM. While convex optimization algorithms enjoy nice theoretical guarantees on both optimization and statistical rates, due to the nuclear norm penalty, they involve a singular value decomposition (SVD) for computing the proximal mapping of nuclear norm at each iteration. The time complexity of SVD is $O(d^3)$ (Golub and Van Loan, 2012), which is computationally prohibitive when the dimension $d$ is extremely high.

Another line of research related to ours is matrix factorization, which has been widely used in practice due to its superior empirical performance. Very recently, algorithms based on alternating minimization and gradient descent have been analyzed for low-rank matrix estimation (Jain et al., 2013; Hardt, 2014; Zhao et al., 2015; Zheng and Lafferty, 2015; Chen and Wainwright, 2015; Tu et al., 2015; Bhojanapalli et al., 2015). However, these work is limited to low-rank matrix estimation, and extending them to low-rank and sparse matrix estimation as in LVGGM turns out to be highly nontrivial. The most related work to ours includes Gu et al. (2016) and Yi et al. (2016), which studied nonconvex optimization for low-rank plus sparse matrix estimation. However, they are limited to robust PCA (Candès et al., 2011) and multi-task regression (Agarwal et al., 2012b) in the noiseless setting. The log determinant structure in LVGGM is substantially more challenging to manipulate than the squared loss function used in robust PCA and multi-task regression, and therefore our proof technique is also different from theirs.

The last but not least line of related work is expectation maximization (EM) algorithm (Balakrishnan et al., 2014; Wang et al., 2014), which shares a similar bivariate structure as our estimator. However, the proof technique used in Balakrishnan et al. (2014); Wang et al. (2014) is not directly applicable to our algorithm, due to the matrix factorization structure in our estimator. Moreover, to overcome the dependency issue between consecutive iterations in the proof, Balakrishnan et al. (2014); Wang et al. (2014) employed sample splitting strategy (Jain et al., 2013; Hardt, 2014), i.e., dividing the whole dataset into $T$ pieces and using a fresh piece of data in each iteration. Unfortunately, the sample splitting technique results in a suboptimal statistical rate, incurring an extra factor of $\sqrt{T}$ in the rate. In sharp contrast, our proof technique does not rely on sample splitting, because we are able to prove a uniform convergence result over a small neighborhood of the unknown parameters, which directly resolves the dependency issue.

## 3 The Proposed Estimator and Algorithm

In this section, we present a new estimator for LVGGM estimation, followed by a new algorithm to solve it.

### 3.1 Latent Variable GGMs

Let $\boldsymbol{X}_O$ be the $d$-dimensional random vector with observed variables and $\boldsymbol{X}_L$ be the $r$-dimensional random vector with latent variables. We assume that the concatenated random vector $\boldsymbol{X} = (\boldsymbol{X}_O^\top, \boldsymbol{X}_L^\top)^\top$ follows a multivariate Gaussian distribution with covariance matrix $\widetilde{\boldsymbol{\Sigma}}$ and sparse precision matrix $\widetilde{\boldsymbol{\Omega}} = \widetilde{\boldsymbol{\Sigma}}^{-1}$. It is well-known that the observed data $\boldsymbol{X}_O$ follows a normal distribution with marginal covariance matrix $\boldsymbol{\Sigma}^* = \widetilde{\boldsymbol{\Sigma}}_{OO}$, which is the top-left block matrix in $\widetilde{\boldsymbol{\Sigma}}$ corresponding to $\boldsymbol{X}_O$. The precision matrix of $\boldsymbol{X}_O$ is then given by Schur complement (Golub and Van Loan, 2012):

$$\boldsymbol{\Omega}^* = (\widetilde{\boldsymbol{\Sigma}}_{OO})^{-1} = \widetilde{\boldsymbol{\Omega}}_{OO} - \widetilde{\boldsymbol{\Omega}}_{OL}\widetilde{\boldsymbol{\Omega}}_{LL}^{-1}\widetilde{\boldsymbol{\Omega}}_{LO}. \tag{3.1}$$

Since we can only observe $\boldsymbol{X}_O$, the marginal precision matrix $\boldsymbol{\Omega}^*$ is generally not sparse. We define $\mathbf{S}^* := \widetilde{\boldsymbol{\Omega}}_{OO}$ and $\mathbf{L}^* := -\widetilde{\boldsymbol{\Omega}}_{OL}\widetilde{\boldsymbol{\Omega}}_{LL}^{-1}\widetilde{\boldsymbol{\Omega}}_{LO}$. Then $\mathbf{S}^*$ is sparse due to the sparsity of $\widetilde{\boldsymbol{\Omega}}$. We do not impose any dependency restriction on $\boldsymbol{X}_O$ and $\boldsymbol{X}_L$, and thus the matrices $\widetilde{\boldsymbol{\Omega}}_{OL}$ and $\widetilde{\boldsymbol{\Omega}}_{LO}$ can be potentially dense. We assume that the number of latent variables is smaller than that of the observed. Therefore, $\mathbf{L}^*$ is low-rank and may be dense. In other words, the precision matrix of LVGGM can be written as

$$\boldsymbol{\Omega}^* = \mathbf{S}^* + \mathbf{L}^*, \tag{3.2}$$

where $\|\mathbf{S}^*\|_{0,0} = s^*$ and $\text{rank}(\mathbf{L}^*) = r$. We refer to Chandrasekaran et al. (2010) for a detailed discussion of LVGGM. It is also worth noting that our estimator and algorithm that are going to be proposed are applicable to any Gaussian graphical model whose precision matrix satisfies (3.2), without necessarily being a LVGGM.

## 3.2 The Proposed Estimator

Suppose that we observe i.i.d. samples $\boldsymbol{X}_1, \ldots, \boldsymbol{X}_n$ from $N(\boldsymbol{0}, \boldsymbol{\Sigma}^*)$. Our goal is to estimate the sparse component $\mathbf{S}^*$ and the low-rank component $\mathbf{L}^*$ of the unknown precision matrix $\boldsymbol{\Omega}^*$ in (3.2). The negative log-likelihood of the Gaussian graphical model is proportional to the following sample loss function up to a constant

$$p_n(\mathbf{S}, \mathbf{L}) = \text{tr}\big[\widehat{\boldsymbol{\Sigma}}(\mathbf{S} + \mathbf{L})\big] - \log |\mathbf{S} + \mathbf{L}|, \tag{3.3}$$

where $\widehat{\boldsymbol{\Sigma}} = 1/n \sum_{i=1}^n \boldsymbol{X}_i \boldsymbol{X}_i^\top$ is the sample covariance matrix, and $|\mathbf{S} + \mathbf{L}|$ is the determinant of $\boldsymbol{\Omega} = \mathbf{S} + \mathbf{L}$. We employ the maximum likelihood principle to estimate $\mathbf{S}^*$ and $\mathbf{L}^*$, which is equivalent to minimizing the negative log-likelihood in (3.3).

The low-rank structure of the precision matrix, i.e., $\mathbf{L}$ poses a great challenge for computation. A typical way is to use nuclear-norm regularized estimator, or rank constrained estimator to estimate $\mathbf{L}$. However, such kind of estimators involves singular value decomposition at each iteration, which is computationally very expensive. To overcome this computational obstacle, we reparameterize $\mathbf{L}$ as the outer product of smaller matrices. More specifically, due to the symmetry of $\mathbf{L}$, it can be reparameterized by $\mathbf{L} = \mathbf{Z}\mathbf{Z}^\top$, where $\mathbf{Z} \in \mathbb{R}^{d \times r}$ and $r > 0$ is the number of latent variables. This kind of reparametrization has recently been used in low-rank matrix estimation (Jain et al., 2013; Hardt, 2014; Zhao et al., 2015; Zheng and Lafferty, 2015; Chen and Wainwright, 2015; Tu et al., 2015) based on matrix factorization. Then we can rewrite the sample loss function in (3.3) to be the following objective function

$$q_n(\mathbf{S}, \mathbf{Z}) = \text{tr}\big[\widehat{\boldsymbol{\Sigma}}(\mathbf{S} + \mathbf{Z}\mathbf{Z}^\top)\big] - \log |\mathbf{S} + \mathbf{Z}\mathbf{Z}^\top|. \tag{3.4}$$

Based on (3.4), we propose a nonconvex estimator using sparsity constrained maximum likelihood estimation:

$$\min_{\mathbf{S}, \mathbf{Z}} \quad q_n(\mathbf{S}, \mathbf{Z}) \quad \text{subject to } \|\mathbf{S}\|_{0,0} \leq s, \tag{3.5}$$

where $s$ is a tuning parameter that controls the sparsity of $\mathbf{S}$, and needs to be larger than $s^*$ in theory.

## 3.3 The Proposed Algorithm

Due to the matrix factorization based reparameterization $\mathbf{L} = \mathbf{Z}\mathbf{Z}^\top$, the objective function in (3.5) is nonconvex. In addition, the sparsity based constraint in (3.5) is nonconvex as well. Therefore, the estimation in (3.5) is essentially a nonconvex optimization problem. We propose to solve it by alternately performing gradient descent with respect to one parameter matrix with the other one fixed. The detailed algorithm is displayed in Algorithm 1.

To explain our alternating minimization algorithm in detail, in each iteration, we first estimate $\mathbf{S}$ while fixing $\mathbf{Z}$, and then switch to estimate $\mathbf{Z}$ while fixing $\mathbf{S}$. Instead of solving each subproblem exactly, we propose to perform one-step gradient descent for $\mathbf{S}$ and $\mathbf{Z}$ alternately, using step sizes $\eta$ and $\eta'$. In Lines 3 and 5 of Algorithm 1, $\nabla_\mathbf{S} q_n(\mathbf{S}, \mathbf{Z})$ and $\nabla_\mathbf{Z} q_n(\mathbf{S}, \mathbf{Z})$ denote the partial gradient of $q_n(\mathbf{S}, \mathbf{Z})$ with respect to $\mathbf{S}$ and $\mathbf{Z}$ respectively. The choice of the step sizes will be clear according to our theory. In practice, one can also use line search to choose the step sizes. Algorithm 1 does not involve singular value decomposition in each iteration, neither solve an exact optimization problem, which makes it much faster than the convex relaxation based algorithms (Chandrasekaran et al.,

**Algorithm 1** Alternating Thresholded Gradient Descent (AltGD) for LVGGM
1: **Input:** function $q_n(\mathbf{S}, \mathbf{Z})$, max number of iterations $T$, $\widehat{\mathbf{S}}^{(0)}, \widehat{\mathbf{Z}}^{(0)}$ generated by Algorithm 2, $\eta, \eta'$, $t = 0$.
2: **repeat**
3:    $\widehat{\mathbf{S}}^{(t+0.5)} = \widehat{\mathbf{S}}^{(t)} - \eta \nabla_{\mathbf{S}} q_n(\widehat{\mathbf{S}}^{(t)}, \widehat{\mathbf{Z}}^{(t)})$;
4:    $\widehat{\mathbf{S}}^{(t+1)} = \mathcal{HT}_s(\widehat{\mathbf{S}}^{(t+0.5)})$, which preserves the $s$ largest magnitudes of $\widehat{\mathbf{S}}^{(t+0.5)}$;
5:    $\widehat{\mathbf{Z}}^{(t+1)} = \widehat{\mathbf{Z}}^{(t)} - \eta' \nabla_{\mathbf{Z}} q_n(\widehat{\mathbf{S}}^{(t)}, \widehat{\mathbf{Z}}^{(t)})$;
6:    $t = t + 1$.
7: **until** $t = T + 1$.
8: **output:** $\widehat{\mathbf{S}}^{(T)}, \widehat{\mathbf{Z}}^{(T)}$.

**Algorithm 2** Initialization
1: **Input:** i.i.d. samples $\mathbf{X}_1, \ldots, \mathbf{X}_n$ from latent variable GGM.
2: $\widehat{\mathbf{\Sigma}} = \frac{1}{n} \sum_{i=1}^n \mathbf{X}_i \mathbf{X}_i^\top$.
3: $\widehat{\mathbf{S}}^{(0)} = \mathcal{HT}_s(\widehat{\mathbf{\Sigma}}^{-1})$, which preserves the $s$ largest magnitudes of $\widehat{\mathbf{\Sigma}}^{-1}$.
4: Compute SVD: $\widehat{\mathbf{\Sigma}}^{-1} - \widehat{\mathbf{S}}^{(0)} = \mathbf{U}\mathbf{D}\mathbf{U}^\top$, where $\mathbf{D}$ is a diagonal matrix. Let $\widehat{\mathbf{Z}}^{(0)} = \mathbf{U}\mathbf{D}_r^{1/2}$, where $\mathbf{D}_r$ is the first $r$ columns of $\mathbf{D}$.
5: **output:** $\widehat{\mathbf{S}}^{(0)}, \widehat{\mathbf{Z}}^{(0)}$.

2010; Meng et al., 2014). The computational overhead of Algorithm 1 mainly comes from the calculation of the partial gradient with respect to $\mathbf{Z}$, whose time complexity is $O(rd^2)$. Therefore, our algorithm has a per-iteration complexity of $O(rd^2)$.

Due to the sparsity constraint on $\mathbf{S}$, i.e., $\|\mathbf{S}\|_{0,0} \leq s$, we apply hard thresholding (Blumensath and Davies, 2009) right after the gradient descent step for $\mathbf{S}$, in Line 4 of Algorithm 1. For a matrix $\mathbf{S} \in \mathbb{R}^{d \times d}$ and an integer $0 \leq s \leq d^2$, the hard thresholding operator $\mathcal{HT}_s(\mathbf{S})$ preserves the $s$ largest magnitudes in $\mathbf{S}$ and sets the rest entries to zero.

As will be seen in our theory, Algorithm 1 is guaranteed to converge to the unknown parameters with a local linear rate, provided that the initial points $\widehat{\mathbf{S}}^{(0)}, \widehat{\mathbf{Z}}^{(0)}$ fall in the small neighborhood of $\mathbf{S}^*$ and $\mathbf{Z}^*$ respectively. In order to find such initial points, we propose an initialization algorithm in Algorithm 2.

By using Algorithms 1 and 2 together, our method enjoys the same theoretical guarantee as convex-relaxation based methods, while it is orders of magnitude faster.

## 4 Main Theory

We present our main theory in this section, which characterizes the convergence rate of Algorithm 1, and the statistical rate of its output. We begin with some definitions and assumptions, which are necessary for our theoretical analysis.

**Assumption 4.1.** There is a constant $\nu > 0$ such that $0 < 1/\nu \leq \lambda_{\min}(\mathbf{\Sigma}^*) \leq \lambda_{\max}(\mathbf{\Sigma}^*) \leq \nu < \infty$, where $\lambda_{\min}(\mathbf{\Sigma}^*)$ and $\lambda_{\max}(\mathbf{\Sigma}^*)$ are the minimal and maximal eigenvalues of $\mathbf{\Sigma}^*$ respectively.

Assumption 4.1 requires the eigenvalues of true covariance matrix $\mathbf{\Sigma}^*$ to be finite and bounded below from zero, which is widely imposed in the literature of Gaussian graphical models (Rothman et al., 2008; Liu et al., 2009; Ravikumar et al., 2011). And the relation between the covariance matrix and the precision matrix $\boldsymbol{\Omega}^* = (\boldsymbol{\Sigma}^*)^{-1}$ immediately yields $1/\nu \leq \lambda_{\min}(\boldsymbol{\Omega}^*) \leq \lambda_{\max}(\boldsymbol{\Omega}^*) \leq \nu$.

It is well understood that the estimation problem of the decomposition $\boldsymbol{\Omega}^* = \mathbf{S}^* + \mathbf{L}^*$ can be ill-posed, where identifiability issue arises when the low-rank matrix $\mathbf{L}^*$ is also sparse (Chandrasekaran et al., 2011; Candes and Recht, 2012). The concept of *incoherence* condition, which was originally proposed for matrix completion (Candes and Recht, 2012), has been adopted in Chandrasekaran et al. (2010, 2011), which ensures the low-rank matrix not to be too sparse by restricting the degree of coherence between singular vectors and the standard basis. Later work such as Agarwal et al. (2012b); Negahban and Wainwright (2012) relaxed this condition to a constraint on the *spikiness ratio*, and showed that *spikiness* condition is milder than incoherence condition. In our theory, we use the notion of *spikiness* as follows.

**Assumption 4.2** (Spikiness Condition (Negahban and Wainwright, 2012))**.** For a matrix $\mathbf{L} \in \mathbb{R}^{d \times d}$, the *spikiness ratio* is defined as $\alpha_{sp}(\mathbf{L}) := d\|\mathbf{L}\|_{\infty,\infty}/\|\mathbf{L}\|_F$. For the low-rank matrix $\mathbf{L}^*$ in (3.2), we assume that there exists a constant $\alpha^* > 0$ such that

$$\|\mathbf{L}^*\|_{\infty,\infty} = \frac{\alpha_{sp}(\mathbf{L}^*) \cdot \|\mathbf{L}^*\|_F}{d} \leq \frac{\alpha^*}{d}. \tag{4.1}$$

Since $\text{rank}(\mathbf{L}^*) = r$, we define $\sigma_{\max} = \sigma_1(\mathbf{L}^*) > 0$ and $\sigma_{\min} = \sigma_r(\mathbf{L}^*) > 0$ to be the maximal and minimal nonzero singular value of $\mathbf{L}^*$ respectively. We observe that the decomposition of low-rank matrix $\mathbf{L}^*$ in Section 3.2 is not unique, since we have $\mathbf{L}^* = (\mathbf{Z}^*\mathbf{U})(\mathbf{Z}^*\mathbf{U})^\top$ for any $r \times r$ orthogonal matrix $\mathbf{U}$. Thus, we define the following solution set for $\mathbf{Z}$:

$$\mathcal{U} = \big\{\widetilde{\mathbf{Z}} \in \mathbb{R}^{d \times r} | \widetilde{\mathbf{Z}} = \mathbf{Z}^*\mathbf{U} \text{ for some } \mathbf{U} \in \mathbb{R}^{r \times r} \text{ with } \mathbf{U}\mathbf{U}^\top = \mathbf{U}^\top\mathbf{U} = \mathbf{I}_r\big\}. \tag{4.2}$$

Note that $\sigma_1(\widetilde{\mathbf{Z}}) = \sqrt{\sigma_{\max}}$ and $\sigma_r(\widetilde{\mathbf{Z}}) = \sqrt{\sigma_{\min}}$ for any $\widetilde{\mathbf{Z}} \in \mathcal{U}$.

To measure the closeness between our estimator for $\mathbf{Z}$ and the unknown parameter $\mathbf{Z}^*$, we use the following distance $d(\cdot, \cdot)$, which is invariant to rotation. Similar definition has been used in Zheng and Lafferty (2015); Tu et al. (2015); Yi et al. (2016).

**Definition 4.3.** Define the distance between $\mathbf{Z}$ and $\mathbf{Z}^*$ as

$$d(\mathbf{Z}, \mathbf{Z}^*) = \min_{\widetilde{\mathbf{Z}} \in \mathcal{U}} \|\mathbf{Z} - \widetilde{\mathbf{Z}}\|_F,$$

where $\mathcal{U}$ is the solution set defined in (4.2).

In the analysis of the linear convergence of Algorithm 1, we require that the initial points lie in small neighborhoods of the unknown parameters. We define two balls around $\mathbf{S}^*$ and $\mathbf{Z}^*$ respectively: $\mathbb{B}_F(\mathbf{S}^*, R) = \{\mathbf{S} \in \mathbb{R}^{d \times d} : \|\mathbf{S} - \mathbf{S}^*\|_F \leq R\}$, $\mathbb{B}_d(\mathbf{Z}^*, R) = \{\mathbf{Z} \in \mathbb{R}^{d \times r} : d(\mathbf{Z}, \mathbf{Z}^*) \leq R\}$.

At the core of our proof technique is a so-called first-order stability condition on the population loss function. In detail, the population loss function is defined as the expectation of sample loss function in (3.3):

$$p(\mathbf{S}, \mathbf{L}) = \text{tr}(\boldsymbol{\Sigma}^*(\mathbf{S} + \mathbf{L})) - \log\big|\mathbf{S} + \mathbf{L}\big|. \tag{4.3}$$

And the first-order stability condition is stated as follows.

**Condition 4.4** (First-order Stability). Suppose $\mathbf{S} \in \mathbb{B}_F(\mathbf{S}^*, R), \mathbf{Z} \in \mathbb{B}_d(\mathbf{Z}^*, R)$ for some $R > 0$; by definition we have $\mathbf{L} = \mathbf{Z}\mathbf{Z}^\top$ and $\mathbf{L}^* = \mathbf{Z}^*\mathbf{Z}^{*\top}$. The gradient of population loss function with respect to $\mathbf{S}$ satisfies

$$\left\|\nabla_\mathbf{S} p(\mathbf{S}, \mathbf{L}) - \nabla_\mathbf{S} p(\mathbf{S}, \mathbf{L}^*)\right\|_F \leq \gamma_2 \cdot \|\mathbf{L} - \mathbf{L}^*\|_F.$$

The gradient of the population loss function with respect to $\mathbf{L}$ satisfies

$$\left\|\nabla_\mathbf{L} p(\mathbf{S}, \mathbf{L}) - \nabla_\mathbf{L} p(\mathbf{S}^*, \mathbf{L})\right\|_F \leq \gamma_1 \cdot \|\mathbf{S} - \mathbf{S}^*\|_F,$$

where $\gamma_1, \gamma_2 > 0$ are constants.

Condition 4.4 requires the population loss function has a variant of Lipschitz continuity for the gradient. Note that the gradient is taken with respect to one variable ($\mathbf{S}$ or $\mathbf{L}$), while the Lipschitz continuity is with respect to the other variable. Also, the Lipschitz property is defined only between the true parameters $\mathbf{S}^*, \mathbf{L}^*$ and arbitrary elements $\mathbf{S} \in \mathbb{B}_F(\mathbf{S}^*, R)$ and $\mathbf{L} = \mathbf{Z}\mathbf{Z}^\top$ such that $\mathbf{Z} \in \mathbb{B}_d(\mathbf{Z}^*, R)$. It should be noted that Condition 4.4, as is verified in the appendix, is inspired by a similar condition originally introduced in (Balakrishnan et al., 2014). We extend it to the loss function of LVGMM with both sparse and low-rank structures, which plays an important role in the analysis.

Now we present our main theory, which characterizes the convergence rate and the statistical error of Algorithm 1.

**Theorem 4.5.** Suppose Assumption 4.1 and Condition 4.4 hold. Let $R = \min\{1/4\sqrt{\sigma_{\max}}, 1/(2\nu), \sqrt{\sigma_{\min}}/(6.5\nu^2)\}$. Suppose the initial solutions satisfy $\widehat{\mathbf{S}}^{(0)} \in \mathbb{B}_F(\mathbf{S}^*, R)$ and $\widehat{\mathbf{Z}}^{(0)} \in \mathbb{B}_d(\mathbf{Z}^*, R)$. In Algorithm 1, let the step sizes satisfy $\eta \leq C_0/(\sigma_{\max}\nu^2)$ and $\eta' \leq C_0\sigma_{\min}/(\sigma_{\max}\nu^4)$, and the sparsity parameter satisfies $s \geq \left(4(1/(2\sqrt{\rho}) - 1)^2 + 1\right)s^*$, where $C_0 > 0$ is a sufficiently small constant. Let $\rho$ and $\tau$ be

$$\rho = \max\left\{1 - \frac{C_1}{\sigma_{\max}\nu^4}, 1 - \frac{C_2\sigma_{\min}^2}{\sigma_{\max}\nu^6}\right\},$$

$$\tau = \max\left\{\frac{C_3}{\sigma_{\max}^2\nu^4}\frac{s^*\log d}{n}, \frac{C_4\sigma_{\min}^2}{\sigma_{\max}\nu^6}\frac{rd}{n}\right\}.$$

Then for any $t \geq 1$, with probability at least $1 - C_5/d$, the output of Algorithm 1 satisfies

$$\max\left\{\left\|\widehat{\mathbf{S}}^{(t+1)} - \mathbf{S}^*\right\|_F^2, d^2(\widehat{\mathbf{Z}}^{(t+1)}, \mathbf{Z}^*)\right\} \leq \frac{\tau}{1 - \sqrt{\rho}} + \sqrt{\rho^{t+1}} \cdot R, \tag{4.4}$$

where $\{C_i\}_{i=1}^5$ are absolute constants.

Theorem 4.5 suggests that the estimation error consists of two terms: the first term is the statistical error, and the second term is the optimization error of our algorithm. We make the following remarks.

**Remark 4.6.** While the derived error bound in (4.4) is for $\widehat{\mathbf{Z}}^{(t)}$, it is in the same order of the error bound for $\widehat{\mathbf{L}}^{(t)}$ by their relation. The statistical error of the output of Algorithm 1 is $\max\left\{O_p(\sqrt{s^*\log d/n}), O_p(\sqrt{rd/n})\right\}$, where $O_p(\sqrt{s^*\log d/n})$ corresponds to the statistical error of $\mathbf{S}^*$, and $O_p(\sqrt{rd/n})$ corresponds to the statistical error of $\mathbf{L}^*$. This matches the minimax optimal rate of estimation errors in Frobenius norm for LVGGM (Chandrasekaran et al., 2010;

Agarwal et al., 2012a; Meng et al., 2014). For the optimization error, i.e., the second term in (4.4), note that $\sigma_{\max}$ and $\sigma_{\min}$ are constants, for a sufficiently small constant $C_0$, we can always ensure $\rho < 1$, and this establishes the linear convergence rate for Algorithm 1. Actually, after $T \geq \max\{O(\log(\nu^4 n/(s^* \log d)), \log(\nu^6 n/(rd))\}$ iterations, the total estimation error of our algorithm achieves the same order as the statistical error.

**Remark 4.7.** Our statistical rate is sharp, because our theoretical analysis is conducted uniformly over the neighborhood of true parameters $\mathbf{S}^*$ and $\mathbf{Z}^*$, rather than doing sample splitting. This is another big advantage of our approach over existing algorithms which are also built up on first-order stability (Balakrishnan et al., 2014; Wang et al., 2014) but rely on sample splitting technique.

In Theorem 4.5, we assumed that the initial points $\widehat{\mathbf{S}}^{(0)}$ and $\widehat{\mathbf{Z}}^{(0)}$ lie in small neighborhoods of $\mathbf{S}^*$ and $\mathbf{Z}^*$ respectively. This condition can be satisfied by Algorithm 2 as is shown in the following theorem.

**Theorem 4.8.** Suppose that Assumptions 4.1 and 4.2 hold. Choose the thresholding parameter as $s > s^*$ in Algorithm 2. Then with probability at least $1 - C'/d$, the initial points $\widehat{\mathbf{S}}^{(0)}, \widehat{\mathbf{Z}}^{(0)}$ output by Algorithm 2 satisfy

$$\left\|\widehat{\mathbf{S}}^{(0)} - \mathbf{S}^*\right\|_F \leq \sqrt{s^* + s}\left(\frac{\alpha^*}{d} + C\|\mathbf{\Omega}^*\|_{1,1}\nu\sqrt{\frac{\log d}{n}}\right),$$

$$d(\widehat{\mathbf{Z}}^{(0)}, \mathbf{Z}^*) \leq \frac{C\alpha^*\sqrt{r(s^* + s)}}{d\sqrt{\sigma_{\min}}} + \frac{C\nu^3}{\sqrt{\sigma_{\min}}}\sqrt{\frac{rd}{n}} + \frac{C\|\mathbf{\Omega}^*\|_{1,1}\nu}{\sqrt{\sigma_{\min}}}\sqrt{\frac{r(s^* + s)\log d}{n}},$$

where $C, C' > 0$ are absolute constants.

**Remark 4.9.** Theorem 4.8 indicates that, using Algorithm 2, we are able to obtain initial solutions that are sufficiently close to $\mathbf{S}^*$ and $\mathbf{Z}^*$ respectively. In particular, since $s$ and $s^*$ are the same order, when the sample size satisfies $n \geq C\nu r s^* \log d/(R^2 \sigma_{\min})$ and the sparsity of the unknown sparse matrix satisfies

$$s^* \leq d^2 R^2 \sigma_{\min}/(Cr\alpha^{*2}),$$

the initial points $\widehat{\mathbf{S}}^{(0)}$ and $\widehat{\mathbf{Z}}^{(0)}$ are guaranteed to lie in the small balls with radius $R$ specified in Theorem 4.5. That is to say, the unknown sparse matrix cannot be too dense.

## 5 Experiments

In this section, we present numerical results on both synthetic and real datasets to verify the theoretical properties of our algorithm, and compare it with the state-of-the-art methods. Specifically, we compare our method, denoted by **AltGD**, with two convex relaxation based methods for estimating LVGGM: (1) LogdetPPA (Chandrasekaran et al., 2010; Wang et al., 2010) for solving log-determinant semidefinite programs, denoted by **PPA**, and (2) the alternating direction method of multipliers in Ma et al. (2013); Meng et al. (2014), denoted by **ADMM**. The implementation of these two methods were downloaded from the authors' website. All numerical experiments were run in MATLAB R2015b on a laptop with Intel Core I5 2.7 GHz CPU and 8GB of RAM.

## 5.1 Synthetic Data

In the synthetic experiment, we generated data from latent variable GGM defined in Section 3.1. In detail, the dimension of observed data is $d$ and the number of latent variables is $r$. We randomly generated a sparse positive definite matrix $\widetilde{\mathbf{\Omega}} \in \mathbb{R}^{(d+r)\times(d+r)}$, with sparsity $s^* = 0.02 * d^2$. According to (3.1), the sparse component of the precision matrix is $\mathbf{S}^* := \widetilde{\mathbf{\Omega}}_{1:d;1:d}$ and the low-rank component is $\mathbf{L}^* := -\widetilde{\mathbf{\Omega}}_{1:d;(d+1):(d+r)} [\widetilde{\mathbf{\Omega}}_{(d+1):(d+r);(d+1):(d+r)}]^{-1} \widetilde{\mathbf{\Omega}}_{(d+1):(d+r);1:d}$. Then we sampled data $\mathbf{X}_i, \ldots, \mathbf{X}_n$ from multivariate normal distribution $N(\mathbf{0}, (\mathbf{\Omega}^*)^{-1})$, where $\mathbf{\Omega}^* = \mathbf{S}^* + \mathbf{L}^*$ is the true precision matrix.

Table 1: Estimation errors of sparse and low-rank components $\mathbf{S}^*$ and $\mathbf{L}^*$ as well as the true precision matrix $\mathbf{\Omega}^*$ in terms of Frobenius norm on different synthetic datasets. Data were generated from LVGGM and results were reported on 10 replicates in each setting.

| Setting | Method | $\|\widehat{\mathbf{S}}^{(T)} - \mathbf{S}^*\|_F$ | $\|\widehat{\mathbf{L}}^{(T)} - \mathbf{L}^*\|_F$ | $\|\widehat{\mathbf{\Omega}}^{(T)} - \mathbf{\Omega}^*\|_F$ | Time $(s)$ |
|---|---|---|---|---|---|
| $d=100, r=2, n=2000$ | PPA | 0.7335±0.0352 | 0.0170±0.0125 | 0.7350±0.0359 | 1.1610 |
| | ADMM | 0.7521±0.0288 | 0.0224±0.0115 | 0.7563±0.0298 | 1.1120 |
| | AltGD | 0.6241±0.0668 | 0.0113±0.0014 | 0.6236±0.0669 | 0.0250 |
| $d=500, r=5, n=10000$ | PPA | 0.9803±0.0192 | 0.0195±0.0046 | 0.9813±0.0192 | 35.7220 |
| | ADMM | 1.0571±0.0135 | 0.0294±0.0041 | 1.0610±0.0134 | 25.8010 |
| | AltGD | 0.8212±0.0143 | 0.0125±0.0000 | 0.8210±0.0143 | 0.4800 |
| $d=1000, r=8, n=2.5\times 10^4$ | PPA | 1.1620±0.0177 | 0.0224±0.0034 | 1.1639±0.0179 | 356.7360 |
| | ADMM | 1.1867±0.0253 | 0.0356±0.0033 | 1.1869±0.0254 | 156.5550 |
| | AltGD | 0.9016±0.0245 | 0.0167±0.0030 | 0.9021±0.0244 | 7.4740 |
| $d=5000, r=10, n=2\times 10^5$ | PPA | 1.4822±0.0302 | 0.0371±0.0052 | 1.4824±0.0120 | 33522.0200 |
| | ADMM | 1.5010±0.0240 | 0.0442±0.0068 | 1.5012±0.0240 | 21090.7900 |
| | AltGD | 1.3449±0.0073 | 0.0208±0.0014 | 1.3449±0.0084 | 445.6730 |

We conducted experiments in different settings of $d, n, s^*$ and $r$. The step sizes of our method were set as $\eta = c_1/(\sigma_{\max}\nu^2)$ and $\eta' = c_1\sigma_{\min}/(\sigma_{\max}\nu^4)$ according to Theorem 4.5, where $c_1 = 0.25$. The thresholding parameter $s$ is set as $c_2 s^*$, where $c_2 > 1$ was selected by 4-fold cross-validation. The regularization parameters for $\ell_{1,1}$-norm and nuclear norm in **PPA** and **ADMM** were selected by 4-fold cross-validation. Under each setting, we repeatedly generated 10 datasets, and calculated the mean and standard error of the estimation error. We summarize the results over 10 replications in Table 1. Note that our algorithm **AltGD** outputs a slightly better estimator in terms of estimation errors compared with **PPA** and **ADMM**. It should also be noted that they do not differ too much because their statistical rates should be in the same order. To demonstrate the efficiency of our algorithm, we reported the mean CPU time in the last column of Table 1. We observe significant speed-ups brought by our algorithm, which is almost 50 times faster than the convex algorithms. In particular, when the dimension $d$ scales up to several thousands, the computation of SVD in **PPA** and **ADMM** takes enormous time and therefore the computational time of **PPA** and **ADMM** increases dramatically.

We illustrate the convergence rate of our algorithm in Figure 1, where the x-axis is iteration number and y-axis is the estimation errors in Frobenius norm. We can see that our algorithm converges in dozens of iterations, which confirms our theoretical guarantee on linear convergence rate. We plot the overall estimation errors against the scaled statistical errors of $\mathbf{S}^{(T)}$ and $\mathbf{L}^{(T)}$ under different configurations of $d, n, s^*$ and $r$ in Figure 2. According to Theorem 4.5, $\|\widehat{\mathbf{S}}^{(t)} - \mathbf{S}^*\|_F$ and $\|\widehat{\mathbf{L}}^{(t)} - \mathbf{L}^*\|_F$ will converge to the statistical errors as the number of iterations $t$ goes up, which are in the order of $O(\sqrt{s^* \log d/n})$ and $O(\sqrt{rd/n})$ respectively. We can see that the estimation errors grow linearly with the theoretical rate, which validates our theoretical guarantee on the minimax optimal statistical rate.

Table 2: Estimation errors of sparse and low-rank components $\mathbf{S}^*$ and $\mathbf{L}^*$ as well as the true precision matrix $\boldsymbol{\Omega}^*$ in terms of Frobenius norm on different synthetic datasets. Data were generated from multivariate distribution where the precision matrix is the sum of an arbitrary sparse matrix and an arbitrary low-rank matrix. Results were reported on 10 replicates in each setting.

| Setting | Method | $\|\widehat{\mathbf{S}}^{(T)} - \mathbf{S}^*\|_F$ | $\|\widehat{\mathbf{L}}^{(T)} - \mathbf{L}^*\|_F$ | $\|\widehat{\boldsymbol{\Omega}}^{(T)} - \boldsymbol{\Omega}^*\|_F$ | Time (s) |
|---|---|---|---|---|---|
| $d=100, r=2, n=2000$ | PPA | 0.5710±0.0319 | 0.6231±0.0261 | 0.8912±0.0356 | 1.6710 |
| | ADMM | 0.6198±0.0361 | 0.5286±0.0308 | 0.8588±0.0375 | 1.2790 |
| | AltGD | 0.5639±0.0905 | 0.4824±0.0323 | 0.7483±0.0742 | 0.0460 |
| $d=500, r=5, n=10000$ | PPA | 0.8140±0.0157 | 0.7802±0.0104 | 1.1363±0.0131 | 43.8000 |
| | ADMM | 0.8140±0.0157 | 0.7803±0.0104 | 1.1363±0.0131 | 25.8980 |
| | AltGD | 0.6139±0.0198 | 0.7594±0.0111 | 0.9718±0.0146 | 0.8690 |
| $d=1000, r=8, n=2.5\times 10^4$ | PPA | 0.9235±0.0193 | 1.1985±0.0084 | 1.4913±0.0162 | 487.4900 |
| | ADMM | 0.9209±0.0212 | 1.2131±0.0084 | 1.4975±0.0171 | 163.9350 |
| | AltGD | 0.7249±0.0158 | 0.9651±0.0093 | 1.2029±0.0141 | 7.1360 |
| $d=5000, r=10, n=2\times 10^5$ | PPA | 1.1883±0.0091 | 1.0970±0.0022 | 1.3841±0.0083 | 44098.6710 |
| | ADMM | 1.2846±0.0089 | 1.1568±0.0023 | 1.5324±0.0085 | 20393.3650 |
| | AltGD | 1.0681±0.0034 | 1.0685±0.0023 | 1.2068±0.0032 | 287.8630 |

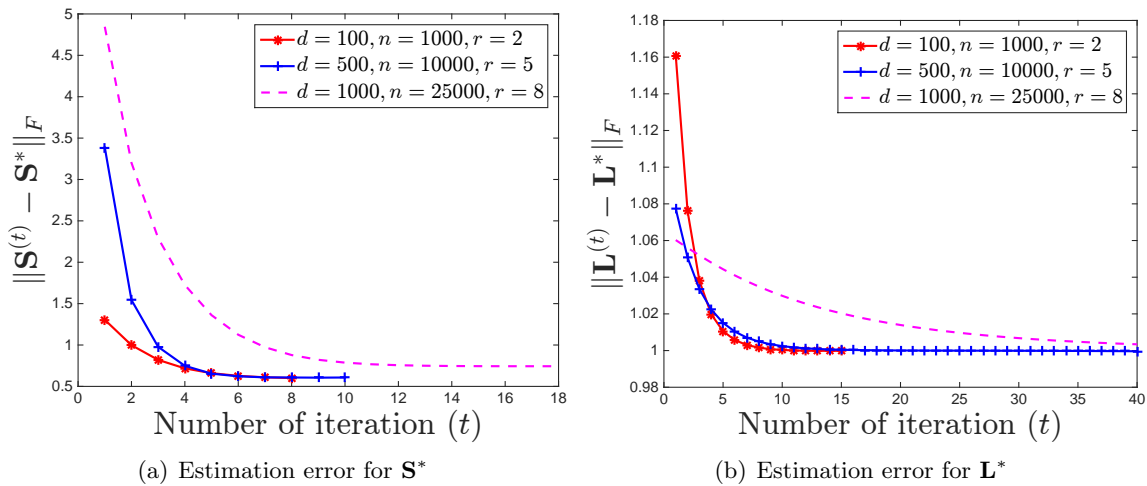

(a) Estimation error for $\mathbf{S}^*$     (b) Estimation error for $\mathbf{L}^*$

Figure 1: Evolution of estimation errors with number of iterations $t$ going up with the sparsity parameter $s^*$ set as $0.02 \times d^2$ and varying $d, n$ and $r$.

In addition, we also conducted experiments on a more general GGM where the precision matrix is the sum of an arbitrary sparse matrix $\mathbf{S}^*$ and an arbitrary low rank matrix $\mathbf{L}^*$. More specifically, $\mathbf{S}^*$ is a symmetric positive definite matrix with entries randomly generated from $[-1, 1]$. $\mathbf{L}^* = \mathbf{Z}^*\mathbf{Z}^{*\top}$, where $\mathbf{Z}^* \in \mathbb{R}^{d\times r}$ with entries randomly generated from $[-1, 1]$. Then we sampled data $\mathbf{X}_i, \ldots, \mathbf{X}_n$ from multivariate normal distribution $N(\mathbf{0}, (\mathbf{\Omega}^*)^{-1})$, where $\mathbf{\Omega}^* = \mathbf{S}^* + \mathbf{L}^*$ is the true precision matrix. Similar to the experiments on LVGGM, we set sparsity $s^* = 0.05 * d^2$, varied the dimension $d$ and number of latent variables $r^*$, and conducted experiments in different settings of $d, n, s^*$ and $r$.

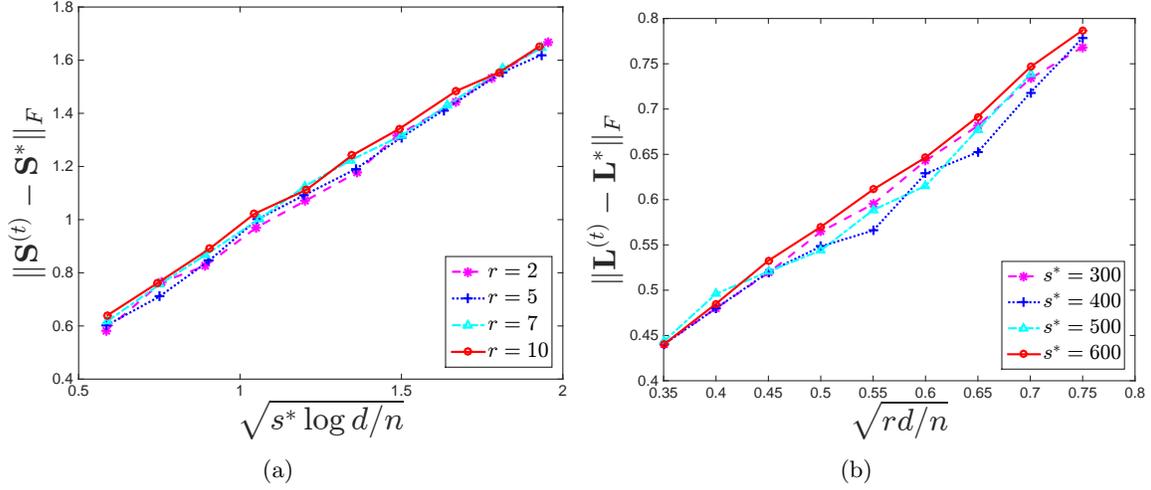

Figure 2: Estimation errors $\|\widehat{\mathbf{S}}^{(T)}-\mathbf{S}^*\|_F$ and $\|\widehat{\mathbf{L}}^{(T)}-\mathbf{L}^*\|_F$ versus scaled statistical errors $\sqrt{s^*\log d/n}$ and $\sqrt{rd/n}$. (a). The estimation error for sparse component $\mathbf{S}^*$, with $r$ fixed and varying $n, d$ and $s^*$. (b). The estimation error for low-rank component $\mathbf{L}^*$ with $s^*$ fixed and varying $n, d$ and $r$.

We repeatedly generated 10 datasets for each setting, and reported the averaged results in Table 2. The parameters for different methods were tuned in the same way as in LVGGM. It can be seen that our method **AltGD** again achieves better estimators in terms of estimation errors in Frobenius norm compared against **PPA** and **ADMM**. Our method **AltGD** is nearly 50 times faster than the other two methods based on convex algorithms.

## 5.2 Real World Data

In this subsection, we apply our method to TCGA breast cancer gene expression data to infer regulatory network. We downloaded the gene expression data from cBioPortal[1]. Here we focused on 299 breast cancer related transcription factors (TFs) and estimated the regulatory relationships for each pair of TFs over two breast cancer subtypes: luminal and basal. We compared our method **AltGD** with **ADMM** and **PPA** which are all based on LVGGM. We also compared with the graphical Lasso (**GLasso**) which only considers the sparse structure of precision matrix and ignores the latent variables; we chose QUIC[2] to solve the GLasso estimator. Regarding the benchmark standard, we used the "regulatory potential scores" between a pair of genes (a TF and a target gene) for these two breast cancer subtypes compiled based on both co-expression and TF ChIP-seq

---

[1] http://www.cbioportal.org/
[2] http://www.cs.utexas.edu/~sustik/QUIC/

binding data from the Cistrome Cancer Database[3].

For luminal subtype, there are $n = 601$ samples and $d = 299$ TFs. The regularization parameters for $\ell_{1,1}$ norm in **GLasso**, for $\ell_{1,1}$ norm and nuclear norm in **PPA** and **ADMM** were tuned by grid search. The step sizes of **AltGD** were set as $\eta = 0.1/\widehat{\nu}^2$ and $\eta' = 0.1/\widehat{\nu}^4$, where $\widehat{\nu}$ is the maximal eigenvalue of sample covariance matrix. The thresholding parameter $s$ and number of latent variables $r$ were tuned by grid search. In Table 3, we present the CPU time of each method. Importantly, we can see that **AltGD** is the fastest among all the methods and is even more than 50 times faster than the second fastest method **ADMM**.

Table 3: Summary of CPU time of different methods on luminal subtype breast cancer dataset.

| Method | GLasso | PPA | ADMM | AltGD |
|---|---|---|---|---|
| Time ($s$) | 38.6310 | 85.0100 | 7.6700 | 0.1500 |

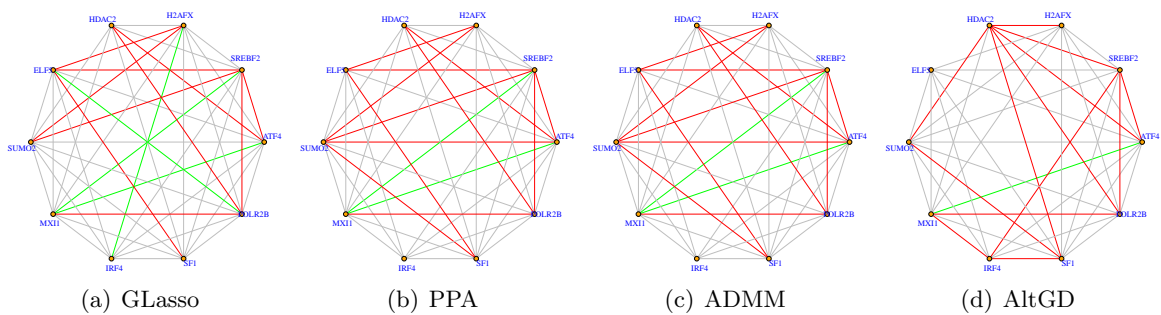

(a) GLasso  (b) PPA  (c) ADMM  (d) AltGD

Figure 3: An example of subnetwork in the transcriptional regulatory network of luminal breast cancer. Here gray edges are the interactions from the Cistrome Cancer Database; red edges are the ones inferred by the respective methods; green edges are incorrectly inferred interactions.

To demonstrate the performances of different methods on recovering the overall transcriptional regulatory network, we randomly selected 10 TFs in benchmark network and plotted the sub-network in Figure 3 which has 70 edges with nonzero regulatory potential scores. Specifically, the gray edges form the benchmark network, the red edges are those identified correctly and the green edges are those incorrectly inferred by each method. We can observe from Figure 3 that the methods based on LVGGM are able to recover more edges accurately than graphical Lasso because of the intervention of latent variables. We remark that all the methods were not able to completely recover the entire regulatory network partly because we only used the gene expression data for TFs (instead of all genes) and the regulatory potential scores from the Cistome Cancer Database also used TF binding information.

To further show the performances of different methods on recovering the edges in the benchmark network that are most related to luminal breast cancer, we chose the top 50 gene pairs with highest regulatory potential scores based on the Cistrome Cancer Database, and plotted the edges identified by each method in Figure 4. Note that the estimated networks of methods based on LVGGM

---
[3] http://cistrome.org/CistromeCancer/

(**ADMM**, **PPA** and **AltGD**) have much more overlaps with the benchmark network on the top 50 edges than **GLasso**, which ignores the latent structure of precision matrix. We also plotted the results for basal subtype breast cancer in Figure 5. We can see that the estimated networks of **ADMM**, **PPA** and **AltGD** again have much more overlaps with the benchmark network on the top 50 edges than **GLasso**, which is consistent with the results for luminal breast cancer.

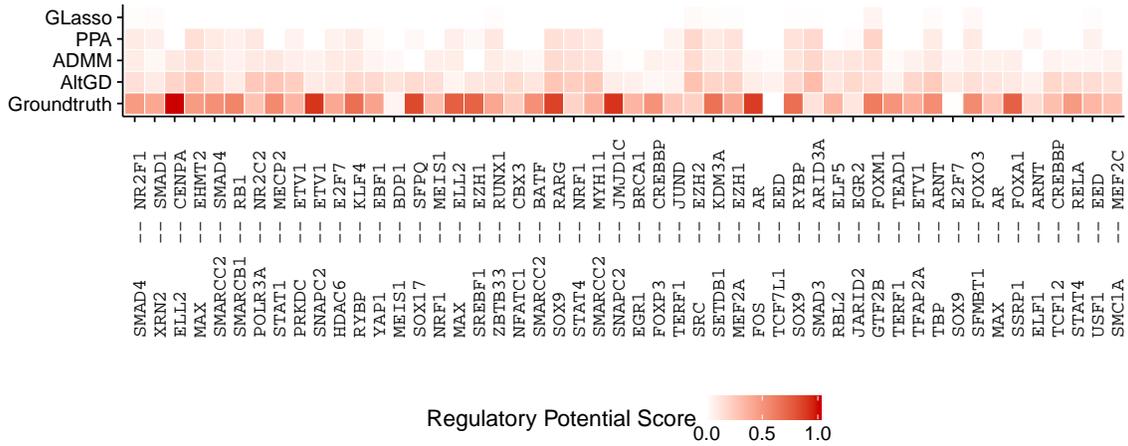

Figure 4: A comparison between the inferred regulatory network as compared to the regulatory potential score from the Cistrome Cancer Database on luminal breast cancer. We chose the top 50 gene pairs in the Database with highest regulatory potential scores.

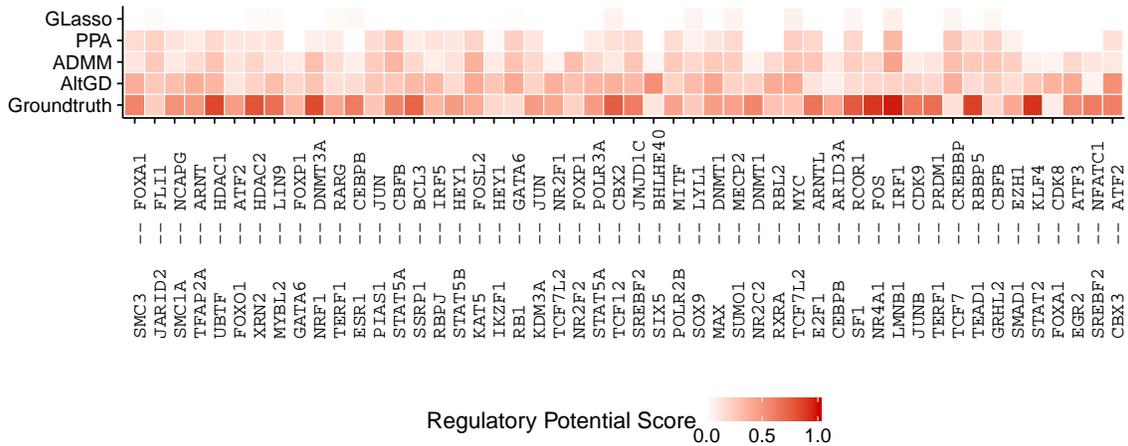

Figure 5: A comparison between the inferred regulatory network as compared to the regulatory potential score from Cistrome Cancer Database on basal breast cancer. We chose the top 50 gene pairs in Cistrome Cancer Database with highest regulatory potential scores.

# 6 Conclusions

In this paper, to speed up the learning of LVGGM, we proposed a sparsity constrained maximum likelihood estimator based on matrix factorization. We developed an efficient alternating gradient descent algorithm, and proved that the proposed algorithm is guaranteed to converge to the unknown sparse and low-rank matrices with a linear convergence rate up to the optimal statical error. Experiments on both synthetic and real world genomic data supported our theory.

# A Proof of Main Theoretical Results

In this section, we prove our main theories.

## A.1 Proof of Theorem 4.5

For simplicity of the proof, we introduce the following notations that give the gradient descent updating based on the population objective function

$$\begin{aligned}
\mathbf{S}^{(t+0.5)} &= \widehat{\mathbf{S}}^{(t)} - \eta \nabla_\mathbf{S} q(\widehat{\mathbf{S}}^{(t)}, \widehat{\mathbf{Z}}^{(t)}), \\
\mathbf{Z}^{(t+1)} &= \widehat{\mathbf{Z}}^{(t)} - \eta' \nabla_\mathbf{Z} q(\widehat{\mathbf{S}}^{(t)}, \widehat{\mathbf{Z}}^{(t)}),
\end{aligned} \quad (A.1)$$

where the population objective function $q(\mathbf{S}, \mathbf{Z}) = \mathbb{E}[q_n(\mathbf{S}, \mathbf{Z})]$ and $q_n(\mathbf{S}, \mathbf{Z})$ is defined in (3.4). Here $\mathbf{S}^{(t+0.5)}$ and $\mathbf{Z}^{(t+1)}$ are the population version of $\widehat{\mathbf{S}}^{(t+0.5)}$ and $\widehat{\mathbf{Z}}^{(t+1)}$ in Algorithm 1. In order to prove our main theorem, we layout some useful lemmas here first.

**Lemma A.1.** Let $\mathbf{S}^{(t+0.5)} = \widehat{\mathbf{S}}^{(t)} - \eta \nabla_\mathbf{S} q(\widehat{\mathbf{S}}^{(t)}, \widehat{\mathbf{Z}}^{(t)})$ be the population version of $\widehat{\mathbf{S}}^{(t+0.5)}$. For the gradient descent updating of $\mathbf{S}$, if step size satisfies $\eta \leq 1/(L+\mu)$, then we have

$$\|\mathbf{S}^{(t+0.5)} - \mathbf{S}^*\|_F^2 \leq \left(1 - \frac{2\eta\mu L}{L+\mu}\right)\|\widehat{\mathbf{S}}^{(t)} - \mathbf{S}^*\|_F^2 + \frac{25\eta^2 \gamma_2^2 \sigma_{\max}}{8} d^2(\widehat{\mathbf{Z}}^{(t)}, \mathbf{Z}^*),$$

where $L = 4\nu^2, \mu = 1/(4\nu^2)$ and $\gamma_2 = 8\nu^2$.

And the corresponding result for $\mathbf{Z}$:

**Lemma A.2.** Let $\mathbf{Z}^{(t+1)} = \widehat{\mathbf{Z}}^{(t)} - \eta' \nabla_\mathbf{Z} q(\widehat{\mathbf{S}}^{(t)}, \widehat{\mathbf{Z}}^{(t)})$ be the population version of $\widehat{\mathbf{Z}}^{(t+1)}$. The gradient descent algorithm of $\mathbf{Z}$ with step size $\eta' \leq 1/[16(L+\mu)\sigma_{\max}]$ satisfies

$$d^2\left(\mathbf{Z}^{(t+1)}, \mathbf{Z}^*\right) \leq \left(1 - \frac{\eta' \sigma_{\min}\mu L}{2(L+\mu)}\right) d^2(\widehat{\mathbf{Z}}^{(t)}, \mathbf{Z}^*) + \frac{25\eta'^2 \gamma_1^2 \sigma_{\max}}{8}\|\widehat{\mathbf{S}}^{(t)} - \mathbf{S}^*\|_F^2,$$

where $L = 4\nu^2, \mu = 1/(4\nu^2)$, and $\gamma_1 = 8\nu^2$. $d(\cdot, \cdot)$ is the distance defined in Definition 4.3.

The following lemma serves similarly as a non-expansive property for hard thresholding operators, which is proved in Lemma 4.1 by Li et al. (2016).

**Lemma A.3** ((Li et al., 2016)). $\boldsymbol{\theta}^* \in \mathbb{R}^d$ is a sparse vector with $\|\boldsymbol{\theta}\|_0 = s^*$. For any $\boldsymbol{\theta} \in \mathbb{R}^d$, let $\mathcal{HT}_s(\cdot)$ be the hard thresholding function which preserves the $s$ largest magnitudes. Then we have

$$\|\mathcal{HT}_s(\boldsymbol{\theta}) - \boldsymbol{\theta}^*\|_2^2 \leq \left(1 + \frac{2\sqrt{s^*}}{\sqrt{s - s^*}}\right)\|\boldsymbol{\theta} - \boldsymbol{\theta}^*\|_2^2.$$

The following lemma gives the statistical error of our model.

**Lemma A.4.** For a given sample with size $n$ and dimension $d$, we use $\epsilon_1(n,d)$ and $\epsilon_2(n,d)$ to denote the statistical errors. More specifically, uniformly for all $\mathbf{S}$ over ball $\mathbb{B}_F(\mathbf{S}^*, R)$, $\mathbf{Z}$ over ball $\mathbb{B}_d(\mathbf{Z}^*, R)$ we have that

$$\|\nabla_{\mathbf{S}} q_n(\mathbf{S}, \mathbf{Z}) - \nabla_{\mathbf{S}} q(\mathbf{S}, \mathbf{Z})\|_{\infty,\infty} \leq \epsilon_1(n,d) = C\sqrt{\frac{\log d}{n}}$$

holds with probability at least $1 - C/d$. And

$$\|\nabla_{\mathbf{Z}} q_n(\mathbf{S}, \mathbf{Z}) - \nabla_{\mathbf{Z}} q(\mathbf{S}, \mathbf{Z})\|_F \leq \epsilon_2(n,d) = C'\nu\sqrt{\sigma_{\max}}\sqrt{\frac{rd}{n}}$$

holds with probability at least $1 - C'/d$.

The above lemma states that the differences between the gradients of the population and sample loss functions with respect to $\mathbf{S}$ and $\mathbf{Z}$ are bounded in terms of different matrix norms. It is pivotal to characterize the statistical error of the estimator from our algorithm.

Now we are going to prove the main theorem.

*Proof of Theorem 4.5.* We show $\widehat{\mathbf{S}}^{(t)} \in \mathbb{B}_F(\mathbf{S}^*, R), \widehat{\mathbf{Z}}^{(t)} \in \mathbb{B}_d(\mathbf{Z}^*, R)$, for all $t = 0, 1, \ldots$ by mathematical induction. We already know the initial points $\widehat{\mathbf{S}}^{(0)} \in \mathbb{B}_F(\mathbf{S}^*, R)$ and $\widehat{\mathbf{Z}}^{(0)} \in \mathbb{B}_d(\mathbf{Z}^*, R)$ by Algorithm 2. Next, suppose that we have $\widehat{\mathbf{S}}^{(t)} \in \mathbb{B}_F(\mathbf{S}^*, R)$, $\widehat{\mathbf{Z}}^{(t)} \in \mathbb{B}_d(\mathbf{Z}^*, R)$ and we want to show this holds for iteration $t+1$ too.

Define $\mathcal{S}^* = \text{supp}(\mathbf{S}^*)$, $\mathcal{S}^{(t)} = \text{supp}(\widehat{\mathbf{S}}^{(t)})$, $\mathcal{S}^{(t+1)} = \text{supp}(\widehat{\mathbf{S}}^{(t+1)})$ and $\bar{\mathcal{S}} = \mathcal{S}^* \cup \mathcal{S}^{(t)} \cup \mathcal{S}^{(t+1)}$. Recall that $\widehat{\mathbf{S}}^{(t+0.5)} = \widehat{\mathbf{S}}^{(t)} + \eta \nabla_{\mathbf{S}} q_n(\widehat{\mathbf{S}}^{(t)}, \widehat{\mathbf{Z}}^{(t)})$ and $\widehat{\mathbf{S}}^{(t+1)}$ preserves the $s$ largest magnitudes in $\widehat{\mathbf{S}}^{(t+0.5)}$, it's easy to verify that

$$\widehat{\mathbf{S}}^{(t+1)} = \mathcal{HT}_s(\widehat{\mathbf{S}}^{(t+0.5)}) = \mathcal{HT}_s\big(\widehat{\mathbf{S}}^{(t)} + \eta\big[\nabla_{\mathbf{S}} q_n(\widehat{\mathbf{S}}^{(t)}, \widehat{\mathbf{Z}}^{(t)})\big]_{\bar{\mathcal{S}}}\big).$$

Thus by Lemma A.3 we have

$$\|\widehat{\mathbf{S}}^{(t+1)} - \mathbf{S}^*\|_F^2 \leq \left(1 + \frac{2\sqrt{s^*}}{\sqrt{s-s^*}}\right)\|\widehat{\mathbf{S}}^{(t)} - \eta\big[\nabla_{\mathbf{S}} q_n(\widehat{\mathbf{S}}^{(t)}, \widehat{\mathbf{Z}}^{(t)})\big]_{\bar{\mathcal{S}}} - \mathbf{S}^*\|_F^2$$

$$\leq 2\left(1 + \frac{2\sqrt{s^*}}{\sqrt{s-s^*}}\right)\|\widehat{\mathbf{S}}^{(t)} - \eta\big[\nabla_{\mathbf{S}} q(\widehat{\mathbf{S}}^{(t)}, \widehat{\mathbf{Z}}^{(t)})\big]_{\bar{\mathcal{S}}} - \mathbf{S}^*\|_F^2$$

$$+ 2\eta^2\left(1 + \frac{2\sqrt{s^*}}{\sqrt{s-s^*}}\right)\|\big[\nabla_{\mathbf{S}} q_n(\widehat{\mathbf{S}}^{(t)}, \widehat{\mathbf{Z}}^{(t)}) - \nabla_{\mathbf{S}} q(\widehat{\mathbf{S}}^{(t)}, \widehat{\mathbf{Z}}^{(t)})\big]_{\bar{\mathcal{S}}}\|_F^2. \quad \text{(A.2)}$$

Note that $|\bar{\mathcal{S}}| \leq s^* + 2s$ and by Lemma A.4, we have with probability at least $1 - C/d$ that

$$\|\big[\nabla_{\mathbf{S}} q_n(\widehat{\mathbf{S}}^{(t)}, \widehat{\mathbf{Z}}^{(t)}) - \nabla_{\mathbf{S}} q(\widehat{\mathbf{S}}^{(t)}, \widehat{\mathbf{Z}}^{(t)})\big]_{\bar{\mathcal{S}}}\|_F \leq \sqrt{s^* + 2s}\|\nabla_{\mathbf{S}} q\big(\widehat{\mathbf{S}}^{(t)}, \widehat{\mathbf{Z}}^{(t)}\big) - \nabla_{\mathbf{S}} q_n\big(\widehat{\mathbf{S}}^{(t)}, \widehat{\mathbf{Z}}^{(t)}\big)\|_{\infty,\infty}$$

$$\leq \sqrt{s^* + 2s}\epsilon_1(n,\delta), \quad \text{(A.3)}$$

where $\epsilon_1(n,\delta) = C\sqrt{\log d/n}$. By definition we have $\bar{\mathcal{S}} = \mathcal{S}^* \cup \mathcal{S}^{(t)} \cup \mathcal{S}^{(t+1)}$, which yields

$$\left\|\widehat{\mathbf{S}}^{(t)} - \eta[\nabla_{\mathbf{S}}q(\widehat{\mathbf{S}}^{(t)},\widehat{\mathbf{Z}}^{(t)})]_{\bar{\mathcal{S}}} - \mathbf{S}^*\right\|_F^2 \leq \left\|\widehat{\mathbf{S}}^{(t)} - \eta\nabla_{\mathbf{S}}q(\widehat{\mathbf{S}}^{(t)},\widehat{\mathbf{Z}}^{(t)}) - \mathbf{S}^*\right\|_F^2$$
$$\leq \left(1 - \frac{2\eta\mu L}{L+\mu}\right)\|\widehat{\mathbf{S}}^{(t)} - \mathbf{S}^*\|_F^2 + \frac{25\eta^2\gamma_2^2\sigma_{\max}}{8}d^2(\widehat{\mathbf{Z}}^{(t)},\mathbf{Z}^*),$$
(A.4)

where the second inequality is due to Lemma A.1. Here $L = 4\nu^2, \mu = 1/(4\nu^2)$ and $\gamma_2 = 8\nu^2$. Submitting (A.3) and (A.4) into (A.2), we obtain with probability at least $1 - C/d$ that

$$\left\|\widehat{\mathbf{S}}^{(t+1)} - \mathbf{S}^*\right\|_F^2 \leq 2\left(1 + \frac{2\sqrt{s^*}}{\sqrt{s-s^*}}\right)\left\{\left(1 - \frac{2\eta\mu L}{L+\mu}\right)\|\widehat{\mathbf{S}}^{(t)} - \mathbf{S}^*\|_F^2 + \frac{25\eta^2\gamma_2^2\sigma_{\max}}{8}d^2(\widehat{\mathbf{Z}}^{(t)},\mathbf{Z}^*)\right.$$
$$\left. + \eta^2(s^*+2s)\epsilon_1^2(n,\delta)\right\}.$$
(A.5)

On the other hand, let $\mathbf{Z}^{(t+1)} = \widehat{\mathbf{Z}}^{(t)} - \eta'\nabla_{\mathbf{Z}}q(\widehat{\mathbf{S}}^{(t)},\widehat{\mathbf{Z}}^{(t)})$. We have

$$d^2(\widehat{\mathbf{Z}}^{(t+1)},\mathbf{Z}^*) = \min_{\widetilde{\mathbf{Z}}\in\mathcal{U}}\|\widehat{\mathbf{Z}}^{(t+1)} - \widetilde{\mathbf{Z}}\|_F^2 \leq 2\|\widehat{\mathbf{Z}}^{(t+1)} - \mathbf{Z}^{(t+1)}\|_F^2 + 2\min_{\widetilde{\mathbf{Z}}\in\mathcal{U}}\|\mathbf{Z}^{(t+1)} - \widetilde{\mathbf{Z}}\|_F^2$$
$$= 2\|\widehat{\mathbf{Z}}^{(t+1)} - \mathbf{Z}^{(t+1)}\|_F^2 + 2d^2(\mathbf{Z}^{(t+1)},\mathbf{Z}^*).$$
(A.6)

By Lemma A.2 we have

$$d^2(\mathbf{Z}^{(t+1)},\mathbf{Z}^*) \leq \left(1 - \frac{\eta'\sigma_{\min}\mu L}{2(L+\mu)}\right)d^2(\widehat{\mathbf{Z}}^{(t)},\mathbf{Z}^*) + \frac{25\eta'^2\gamma_1^2\sigma_{\max}}{8}\|\widehat{\mathbf{S}}^{(t)} - \mathbf{S}^*\|_F^2,$$
(A.7)

where $L = 4\nu^2, \mu = 1/(4\nu^2)$, and $\gamma_1 = 8\nu^2$. By Lemma A.4, we have with probability at least $1 - C'/d$ that

$$\left\|\widehat{\mathbf{Z}}^{(t+1)} - \mathbf{Z}^{(t+1)}\right\|_F = \eta'\left\|\nabla_{\mathbf{Z}}q_n(\widehat{\mathbf{S}}^{(t)},\widehat{\mathbf{Z}}^{(t)}) - \nabla_{\mathbf{Z}}q(\widehat{\mathbf{S}}^{(t)},\widehat{\mathbf{Z}}^{(t)})\right\|_F \leq \eta'\epsilon_2(n,\delta),$$
(A.8)

where $\epsilon_2(n,\delta) = C'\nu\sqrt{\sigma_{\max}}\sqrt{rd/n}$. Substituting (A.6) with (A.7) and (A.8), we obtain

$$d^2(\widehat{\mathbf{Z}}^{(t+1)},\mathbf{Z}^*) \leq 2\left(1 - \frac{\eta'\sigma_{\min}\mu L}{2(L+\mu)}\right)d^2(\widehat{\mathbf{Z}}^{(t)},\mathbf{Z}^*) + \frac{25\eta'^2\gamma_1^2\sigma_{\max}}{4}\|\widehat{\mathbf{S}}^{(t)} - \mathbf{S}^*\|_F^2 + 2\eta'^2\epsilon_2^2(n,\delta) \quad (A.9)$$

holds with probability at least $1 - C'/d$. Combining (A.5) and (A.9), we then have

$$\max\left\{\left\|\widehat{\mathbf{S}}^{(t+1)} - \mathbf{S}^*\right\|_F^2, d^2(\widehat{\mathbf{Z}}^{(t+1)},\mathbf{Z}^*)\right\}$$
$$\leq 2\left(1 + \frac{2\sqrt{s^*}}{\sqrt{s-s^*}}\right)\underbrace{\max\left\{1 - \frac{2\eta\mu L}{L+\mu} + \frac{25\eta^2\gamma_2^2\sigma_{\max}}{8}, 1 - \frac{\eta'\sigma_{\min}\mu L}{2(L+\mu)} + \frac{25\eta'^2\gamma_1^2\sigma_{\max}}{8}\right\}}_{\rho}$$
$$\cdot \max\left\{\|\widehat{\mathbf{S}}^{(t)} - \mathbf{S}^*\|_F^2, d^2(\widehat{\mathbf{Z}}^{(t)},\mathbf{Z}^*)\right\}$$
$$+ \underbrace{\max\left\{2\left(1 + \frac{2\sqrt{s^*}}{\sqrt{s-s^*}}\right)\eta^2(s^*+2s)\epsilon_1^2(n,\delta), 2\eta'^2\epsilon_2^2(n,\delta)\right\}}_{\tau} \quad (A.10)$$

holds with probability at least $1-\max\{C, C'\}/d$. Recall that by Lemma A.1 and Lemma A.2 we have $L = 4\nu^2$, $\mu = 1/(4\nu^2)$, $\gamma_1 = 8\nu^2$ and $\gamma_2 = 8\nu^2$. And by Lemma A.4, we have $\epsilon_1(n,\delta) = C\sqrt{\log d/n}$, $\epsilon_2(n,\delta) = C'\nu\sqrt{\sigma_{\max}}\sqrt{rd/n}$. Note that in Lemma A.1 and Lemma A.2, we require the step sizes satisfy $\eta \leq 1/[16(L+\mu)]$ and $\eta' \leq 1/[16(L+\mu)\sigma_{\max}]$. In order to ensure the convergence of our algorithm, we require that $\rho < 1$. Thus we choose $\eta = C_0/(\sigma_{\max}\nu^2)$ and $\eta' = C_0\sigma_{\min}/(\sigma_{\max}\nu^4)$, where $C_0 > 0$ is a sufficient small constant. Then we have

$$\rho = \max\left\{1 - \frac{C_1}{\sigma_{\max}\nu^4}, 1 - \frac{C_2\sigma_{\min}^2}{\sigma_{\max}\nu^6}\right\}, \qquad \tau = \max\left\{\frac{C_3}{\sigma_{\max}^2\nu^4}\frac{s^*\log d}{n}, \frac{C_4\sigma_{\min}^2}{\sigma_{\max}\nu^6}\frac{rd}{n}\right\}, \qquad (A.11)$$

where $C_1, C_2, C_3, C_4 > 0$ are absolute constants. When we choose the thresholding parameter as $s \geq \big(4(1/(2\sqrt{\rho})-1)^2+1\big)s^*$, it's easy to derive $2\big(1+2\sqrt{s^*/(s-s^*)}\big) \leq 1/\sqrt{\rho}$. Then we have

$$\max\left\{\|\widehat{\mathbf{S}}^{(t+1)} - \mathbf{S}^*\|_F^2, d^2(\widehat{\mathbf{Z}}^{(t+1)}, \mathbf{Z}^*)\right\} \leq \sqrt{\rho}\max\left\{\|\widehat{\mathbf{S}}^{(t)} - \mathbf{S}^*\|_F^2, d^2(\widehat{\mathbf{Z}}^{(t)}, \mathbf{Z}^*)\right\} + \tau$$
$$\leq \sqrt{\rho}R^2 + (1-\sqrt{\rho})R^2 = R^2,$$

where in the second inequality we use the fact that when the sample size $n$ is sufficient large, we are able to ensure $\tau \leq (1-\sqrt{\rho})R^2$. Therefore, we have $\widehat{\mathbf{S}}^{(t+1)} \in \mathbb{B}_F(\mathbf{S}^*, R)$ and $\widehat{\mathbf{Z}}^{(t+1)} \in \mathbb{B}_d(\mathbf{Z}^*, R)$. By mathematical induction, we have $\widehat{\mathbf{S}}^{(t)} \in \mathbb{B}_F(\mathbf{S}^*, R)$ and $\widehat{\mathbf{Z}}^{(t)} \in \mathbb{B}_d(\mathbf{Z}^*, R)$, for any $t = 0, 1, \ldots$

Since (A.10) holds uniformly for all $t$, we further obtain with probability at least $1 - C_5$ that

$$\max\left\{\|\widehat{\mathbf{S}}^{(t+1)} - \mathbf{S}^*\|_F^2, d^2(\widehat{\mathbf{Z}}^{(t+1)}, \mathbf{Z}^*)\right\} \leq \sqrt{\rho}\max\left\{\|\widehat{\mathbf{S}}^{(t)} - \mathbf{S}^*\|_F^2, d^2(\widehat{\mathbf{Z}}^{(t)}, \mathbf{Z}^*)\right\} + \tau$$
$$\leq \frac{\tau}{1-\sqrt{\rho}} + \sqrt{\rho^{t+1}} \cdot R,$$

where $\rho$ and $\tau$ are defined in (A.11) and $C_5 = \max\{C, C'\}$ is a positive constant, which completes the proof. $\square$

## A.2 Proof of Theorem 4.8

In this subsection, we prove that our initial points output by Algorithm 2 are in small neighborhoods of $\mathbf{S}^*$ and $\mathbf{Z}^*$. Note that our analysis of the initialization algorithm is inspired by the proof of Theorem 1 in Yi et al. (2016), and extends that to the noisy case. We first lay out the following lemma, which is useful in our proof.

**Lemma A.5.** For any symmetric matrix $\mathbf{A} \in \mathbb{R}^{d \times d}$ with $\|\mathbf{A}\|_{0,0} = s_0$, we have

$$\|\mathbf{A}\|_2 \leq \sqrt{s_0}\|\mathbf{A}\|_{\infty,\infty}.$$

Now we prove Theorem 4.8.

*Proof.* Let $\mathbf{E} = \mathbf{\Omega}^* - \widehat{\mathbf{\Sigma}}^{-1} = \mathbf{S}^* + \mathbf{L}^* - \widehat{\mathbf{\Sigma}}^{-1}$, where $\widehat{\mathbf{\Sigma}} = 1/n\sum_{i=1}^{n} \mathbf{X}_i\mathbf{X}_i^\top$ is the sample covariance matrix. By Algorithm 2, we have $\widehat{\mathbf{S}}^{(0)} = \mathcal{HT}_s(\widehat{\mathbf{\Sigma}}^{-1})$. We define $\mathbf{Y} = \mathbf{\Omega}^* - \widehat{\mathbf{S}}^{(0)} = \mathbf{E} + \widehat{\mathbf{\Sigma}}^{-1} - \widehat{\mathbf{S}}^{(0)}$, which immediately implies that $\mathbf{Y} - \mathbf{L}^* = \mathbf{S}^* - \widehat{\mathbf{S}}^{(0)}$ and that $\mathrm{supp}(\mathbf{Y} - \mathbf{L}^*) = \mathrm{supp}(\widehat{\mathbf{S}}^{(0)}) \cup \mathrm{supp}(\mathbf{S}^*)$. Specifically,

- For $(j,k) \in \mathrm{supp}(\widehat{\mathbf{S}}^{(0)})$, we have $[\mathbf{Y} - \mathbf{L}^*]_{jk} = [\mathbf{E} - \mathbf{L}^*]_{jk}$, since $[\widehat{\mathbf{\Sigma}}^{-1} - \widehat{\mathbf{S}}^{(0)}]_{jk} = 0$ by thresholding.

- For $(j,k) \in \text{supp}(\mathbf{S}^*)/\text{supp}(\widehat{\mathbf{S}}^{(0)})$, we have $|[\mathbf{Y} - \mathbf{L}^*]_{jk}| = |S^*_{jk}| \leq 2\|\mathbf{L}^*\|_{\infty,\infty} + \|\mathbf{E}\|_{\infty,\infty}$. Otherwise $|[\widehat{\boldsymbol{\Sigma}}^{-1}]_{jk}| = |[\mathbf{S}^* + \mathbf{L}^* - \mathbf{E}]_{jk}| \geq |S^*_{jk}| - |[\mathbf{L}^* - \mathbf{E}]_{jk}| \geq \|\mathbf{L}^*\|_{\infty,\infty}$. Since $\|\mathbf{S}^*\|_{0,0} \leq s^*$ and $s \geq s^*$, this means that $|[\widehat{\boldsymbol{\Sigma}}^{-1}]_{jk}|$ is greater than at least $d - s^* \geq d - s$ entries in $\widehat{\boldsymbol{\Sigma}}^{-1}$, which immediately yields that $(j,k) \in \text{supp}(\widehat{\mathbf{S}}^{(0)})$. This contradiction leads to our claim that $|[\mathbf{Y} - \mathbf{L}^*]_{jk}| = |S^*_{jk}| \leq 2\|\mathbf{L}^*\|_{\infty,\infty} + \|\mathbf{E}\|_{\infty,\infty}$.

Thus, we've proved that

$$\|\mathbf{Y} - \mathbf{L}^*\|_{\infty,\infty} \leq 2\|\mathbf{L}^*\|_{\infty,\infty} + \|\mathbf{E}\|_{\infty,\infty}. \tag{A.12}$$

For $\mathbf{L}^* = \mathbf{V}^*\mathbf{D}^*\mathbf{V}^{*\top}$, by spikiness condition of $\mathbf{L}^*$ in Assumption 4.2, we have

$$\|\mathbf{L}^*\|_{\infty,\infty} \leq \frac{\alpha^*}{d}. \tag{A.13}$$

Moreover, since $\mathbf{E} = \boldsymbol{\Sigma}^{*-1} - \widehat{\boldsymbol{\Sigma}}^{-1}$, we notice that

$$\|\boldsymbol{\Sigma}^{*-1} - \widehat{\boldsymbol{\Sigma}}^{-1}\|_{\infty,\infty} = \|\widehat{\boldsymbol{\Sigma}}^{-1}(\widehat{\boldsymbol{\Sigma}} - \boldsymbol{\Sigma}^*)\boldsymbol{\Sigma}^{*-1}\|_{\infty,\infty} \leq C\|\boldsymbol{\Omega}^*\|_{1,1}\nu\sqrt{\frac{\log d}{n}} \tag{A.14}$$

holds with probability at least $1 - C_0/d$, where the last inequality is due to Lemma D.2. Combining (A.12), (A.13) and (A.14), it finally yields that

$$\|\widehat{\mathbf{S}}^{(0)} - \mathbf{S}^*\|_F \leq \sqrt{s^* + s}\|\mathbf{Y} - \mathbf{L}^*\|_{\infty,\infty} \leq \sqrt{s^* + s}\left(\frac{\alpha^*}{d} + C\|\boldsymbol{\Omega}^*\|_{1,1}\nu\sqrt{\frac{\log d}{n}}\right)$$

holds with probability at least $1 - C_0/d$. It follows from Lemma A.5 that

$$\|\mathbf{Y} - \mathbf{L}^*\|_2 \leq \sqrt{s^* + s}\|\mathbf{Y} - \mathbf{L}^*\|_{\infty,\infty} \leq 2\sqrt{s^* + s}\|\mathbf{L}^*\|_{\infty,\infty} + \sqrt{s^* + s}\|\mathbf{E}\|_{\infty,\infty} \tag{A.15}$$

Since $\widehat{\mathbf{Z}}^{(0)}\widehat{\mathbf{Z}}^{(0)\top}$ is the rank $r$ approximation of $\widehat{\boldsymbol{\Sigma}}^{-1} - \widehat{\mathbf{S}}^{(0)} = \mathbf{Y} - \mathbf{E}$, we have

$$\|\widehat{\mathbf{Z}}^{(0)}\widehat{\mathbf{Z}}^{(0)\top} - (\mathbf{Y} - \mathbf{E})\|_2 = \sigma_{r+1}(\mathbf{Y} - \mathbf{E}).$$

Noting that $\sigma_{r+1}(\mathbf{L}^*) = 0$, applying Weyl's theorem yields

$$|\sigma_{r+1}(\mathbf{Y} - \mathbf{E}) - \sigma_{r+1}(\mathbf{L}^*)| \leq \|(\mathbf{Y} - \mathbf{E}) - \mathbf{L}^*\|_2,$$

which immediately implies

$$\|\widehat{\mathbf{Z}}^{(0)}\widehat{\mathbf{Z}}^{(0)\top} - \mathbf{L}^*\|_2 \leq \|\widehat{\mathbf{Z}}^{(0)}\widehat{\mathbf{Z}}^{(0)\top} - (\mathbf{Y} - \mathbf{E})\|_2 + \|\mathbf{Y} - \mathbf{E} - \mathbf{L}^*\|_2$$
$$\leq 2\|\mathbf{Y} - \mathbf{E} - \mathbf{L}^*\|_2.$$

Thus submitting (A.13), (A.15) and Lemma D.3 into the above inequality, we obtain

$$\|\widehat{\mathbf{Z}}^{(0)}\widehat{\mathbf{Z}}^{(0)\top} - \mathbf{L}^*\|_F \leq 2\sqrt{r}\big(\|\mathbf{Y} - \mathbf{L}^*\|_2 + \|\mathbf{E}\|_2\big)$$
$$\leq 2\sqrt{r(s^* + s)}\|\mathbf{L}^*\|_{\infty,\infty} + 2\sqrt{r(s^* + s)}\|\mathbf{E}\|_{\infty,\infty} + 2\sqrt{r}\|\mathbf{E}\|_2$$
$$\leq \frac{2\alpha^*\sqrt{r(s^* + s)}}{d} + 2C\|\boldsymbol{\Omega}^*\|_{1,1}\nu\sqrt{\frac{r(s^* + s)\log d}{n}} + 2C_1\nu^3\sqrt{\frac{rd}{n}}, \tag{A.16}$$

with probability at least $1 - C'/d$, where $C' = \max\{C_0, C_1\}$. And by Lemma D.4 we further get

$$d(\widehat{\mathbf{Z}}^{(0)}, \mathbf{Z}^*) \leq \frac{C\alpha^*\sqrt{r(s^* + s)}}{\sqrt{\sigma_{\min}}d} + \frac{C\|\boldsymbol{\Omega}^*\|_{1,1}\nu}{\sqrt{\sigma_{\min}}}\sqrt{\frac{r(s^* + s)\log d}{n}} + \frac{C\nu^3}{\sqrt{\sigma_{\min}}}\sqrt{\frac{rd}{n}}, \tag{A.17}$$

with probability at least $1 - C'/d$, which completes the proof. □

# B Proof of Supporting Lemmas

In this section, we prove the lemmas used in the proof of main theorem. We first lay out some useful lemmas. The first lemma is about the strong convexity and smoothness.

**Lemma B.1.** The population loss function $p(\mathbf{S}, \mathbf{L}^*)$ is $\mu$-strongly convex and $L$-smooth with respect to $\mathbf{S}$, namely,

$$\mu \|\mathbf{S} - \mathbf{S}^*\|_F^2 \leq \langle \nabla_\mathbf{S} p(\mathbf{S}, \mathbf{L}^*) - \nabla_\mathbf{S} p(\mathbf{S}^*, \mathbf{L}^*), \mathbf{S} - \mathbf{S}^* \rangle \leq L \|\mathbf{S} - \mathbf{S}^*\|_F^2,$$

for all $\mathbf{S} \in \mathbb{B}_F(\mathbf{S}^*, R)$, where $\mu = 1/(4\nu^2)$ and $L = 4\nu^2$. Similarly, $p(\mathbf{S}^*, \mathbf{L})$ is $\mu$-strongly convex and $L$-smooth with respect to $\mathbf{L}$:

$$\mu \|\mathbf{L} - \mathbf{L}^*\|_F^2 \leq \langle \nabla_\mathbf{L} p(\mathbf{S}^*, \mathbf{L}) - \nabla_\mathbf{L} p(\mathbf{S}^*, \mathbf{L}^*), \mathbf{L} - \mathbf{L}^* \rangle \leq L \|\mathbf{L} - \mathbf{L}^*\|_F^2,$$

for $\mathbf{L} = \mathbf{Z}\mathbf{Z}^\top, \mathbf{L}^* = \mathbf{Z}^*\mathbf{Z}^{*\top}$ and $\mathbf{Z} \in \mathbb{B}_d(\mathbf{Z}^*, R)$. Here we use $\nabla_\mathbf{L} p(\mathbf{S}, \mathbf{L})$ to denote the gradient of the loss function with respect to $\mathbf{L}$.

In the following lemma, we show that the first-order stability, i.e., Condition 4.4 on the population loss function holds for $\mathbf{S}$ and $\mathbf{L}$.

**Lemma B.2.** For all $\mathbf{S} \in \mathbb{B}_F(\mathbf{S}^*, R)$ and $\mathbf{Z} \in \mathbb{B}_d(\mathbf{Z}^*, R)$, by definition we have $\mathbf{L} = \mathbf{Z}\mathbf{Z}^\top$ and $\mathbf{L}^* = \mathbf{Z}^*\mathbf{Z}^{*\top}$. We have the following properties for gradient with respect to $\mathbf{S}$ and $\mathbf{L}$

$$\|\nabla_\mathbf{L} p(\mathbf{S}, \mathbf{L}) - \nabla_\mathbf{L} p(\mathbf{S}^*, \mathbf{L})\|_F \leq \gamma_1 \|\mathbf{S} - \mathbf{S}^*\|_F,$$
$$\|\nabla_\mathbf{S} p(\mathbf{S}, \mathbf{L}) - \nabla_\mathbf{S} p(\mathbf{S}, \mathbf{L}^*)\|_F \leq \gamma_2 \|\mathbf{L} - \mathbf{L}^*\|_F,$$

where $\gamma_1 = \gamma_2 = 8\nu^2$.

## B.1 Proof of Lemma A.1

*Proof.* Since $\mathbf{S}^{(t+0.5)} = \widehat{\mathbf{S}}^{(t)} - \eta \nabla_\mathbf{S} q(\widehat{\mathbf{S}}^{(t)}, \widehat{\mathbf{Z}}^{(t)})$, we have

$$\begin{aligned}
\|\mathbf{S}^{(t+0.5)} - \mathbf{S}^*\|_F^2 &= \|\widehat{\mathbf{S}}^{(t)} - \eta \nabla_\mathbf{S} q(\widehat{\mathbf{S}}^{(t)}, \widehat{\mathbf{Z}}^{(t)}) - \mathbf{S}^*\|_F^2 \\
&\leq \|\widehat{\mathbf{S}}^{(t)} - \mathbf{S}^*\|_F^2 - 2\eta \langle \nabla_\mathbf{S} q(\widehat{\mathbf{S}}^{(t)}, \widehat{\mathbf{Z}}^{(t)}), \widehat{\mathbf{S}}^{(t)} - \mathbf{S}^* \rangle + \eta^2 \|\nabla_\mathbf{S} q(\widehat{\mathbf{S}}^{(t)}, \widehat{\mathbf{Z}}^{(t)})\|_F^2 \\
&= \|\widehat{\mathbf{S}}^{(t)} - \mathbf{S}^*\|_F^2 - 2\eta \underbrace{\langle \nabla_\mathbf{S} q(\widehat{\mathbf{S}}^{(t)}, \widehat{\mathbf{Z}}^{(t)}) - \nabla_\mathbf{S} q(\mathbf{S}^*, \widehat{\mathbf{Z}}^{(t)}), \widehat{\mathbf{S}}^{(t)} - \mathbf{S}^* \rangle}_{I_1} \\
&\quad - 2\eta \underbrace{\langle \nabla_\mathbf{S} q(\mathbf{S}^*, \widehat{\mathbf{Z}}^{(t)}), \widehat{\mathbf{S}}^{(t)} - \mathbf{S}^* \rangle}_{I_2} + \eta^2 \underbrace{\|\nabla_\mathbf{S} q(\widehat{\mathbf{S}}^{(t)}, \widehat{\mathbf{Z}}^{(t)})\|_F^2}_{I_3}.
\end{aligned} \quad (B.1)$$

Since by Lemma B.1 $p(\mathbf{S}, \mathbf{Z}^*\mathbf{Z}^{*\top})$ is $\mu$-strongly convex and $L$-smooth regarding with $\mathbf{S}$ around $\mathbf{S}^*$, and note that $p(\mathbf{S}, \mathbf{Z}^*\mathbf{Z}^{*\top}) = q(\mathbf{S}, \mathbf{Z}^*)$, we also have that $q(\mathbf{S}, \mathbf{Z}^*)$ is $\mu$-strongly convex and $L$-smooth regarding with $\mathbf{S}$ around $\mathbf{S}^*$. For term $I_1$, applying Lemma D.1 yields

$$\begin{aligned}
I_1 &= \langle \nabla_\mathbf{S} q(\widehat{\mathbf{S}}^{(t)}, \widehat{\mathbf{Z}}^{(t)}) - \nabla_\mathbf{S} q(\mathbf{S}^*, \widehat{\mathbf{Z}}^{(t)}), \widehat{\mathbf{S}}^{(t)} - \mathbf{S}^* \rangle \\
&\geq \frac{\mu L}{L + \mu} \|\widehat{\mathbf{S}}^{(t)} - \mathbf{S}^*\|_F^2 + \frac{1}{L + \mu} \|\nabla_\mathbf{S} q(\widehat{\mathbf{S}}^{(t)}, \widehat{\mathbf{Z}}^{(t)}) - \nabla_\mathbf{S} q(\mathbf{S}^*, \widehat{\mathbf{Z}}^{(t)})\|_F^2.
\end{aligned} \quad (B.2)$$

For term $I_2$ in (B.1), noting that $\nabla_\mathbf{S} q(\mathbf{S}^*, \widehat{\mathbf{Z}}^*) = 0$ and the fact that $\nabla_\mathbf{S} q(\mathbf{S}, \mathbf{Z}) = \nabla_\mathbf{\Omega} q(\mathbf{\Omega})$ where $\mathbf{\Omega} = \mathbf{S} + \mathbf{L}$ and $\mathbf{L} = \mathbf{Z}\mathbf{Z}^\top$, we have

$$I_2 = \langle \nabla_\mathbf{S} q(\mathbf{S}^*, \widehat{\mathbf{Z}}^{(t)}) - \nabla_\mathbf{S} q(\mathbf{S}^*, \mathbf{Z}^*), \widehat{\mathbf{S}}^{(t)} - \mathbf{S}^* \rangle = \langle \nabla_\mathbf{\Omega} q(\mathbf{S}^* + \widehat{\mathbf{L}}^{(t)}) - \nabla_\mathbf{\Omega} q(\mathbf{S}^* + \mathbf{L}^*), \widehat{\mathbf{S}}^{(t)} - \mathbf{S}^* \rangle.$$

Applying mean value theorem we further obtain

$$\begin{aligned}
I_2 &= \text{vec}(\widehat{\mathbf{L}}^{(t)} - \mathbf{L}^*)^\top \nabla_\mathbf{\Omega}^2 q(\mathbf{S}^* + (1-t)\mathbf{L}^* + t\widehat{\mathbf{L}}^{(t)}) \text{vec}(\widehat{\mathbf{S}}^{(t)} - \mathbf{S}^*) \\
&\geq \lambda_{\min}(\nabla_\mathbf{\Omega}^2 q(\mathbf{\Omega}^* + t(\widehat{\mathbf{L}}^{(t)} - \mathbf{L}^*))) \|\widehat{\mathbf{L}}^{(t)} - \mathbf{L}^*\|_F \cdot \|\widehat{\mathbf{S}}^{(t)} - \mathbf{S}^*\|_F,
\end{aligned} \tag{B.3}$$

for some $t \in (0, 1)$. Easy calculation and the properties of Kronecker product yield

$$\begin{aligned}
\lambda_{\min}(\nabla_\mathbf{\Omega}^2 q(\mathbf{\Omega}^* + t(\widehat{\mathbf{L}}^{(t)} - \mathbf{L}^*))) &= \lambda_{\min}((\mathbf{\Omega}^* + t(\widehat{\mathbf{L}}^{(t)} - \mathbf{L}^*))^{-1} \otimes (\mathbf{\Omega}^* + t(\widehat{\mathbf{L}}^{(t)} - \mathbf{L}^*))^{-1}) \\
&= (\lambda_{\max}(\mathbf{\Omega}^* + t(\widehat{\mathbf{L}}^{(t)} - \mathbf{L}^*)))^{-2} \\
&\geq (\nu + t\|\widehat{\mathbf{L}}^{(t)} - \mathbf{L}^*\|_2)^{-2} \\
&\geq \frac{1}{4\nu^2}.
\end{aligned} \tag{B.4}$$

Finally, we are going to bound term $I_3$ in (B.1). Specifically, we have

$$\begin{aligned}
\|\nabla_\mathbf{S} q(\widehat{\mathbf{S}}^{(t)}, \widehat{\mathbf{Z}}^{(t)})\|_F^2 &\leq 2\|\nabla_\mathbf{S} q(\widehat{\mathbf{S}}^{(t)}, \widehat{\mathbf{Z}}^{(t)}) - \nabla_\mathbf{S} q(\mathbf{S}^*, \widehat{\mathbf{Z}}^{(t)})\|_F^2 + 2\|\nabla_\mathbf{S} q(\mathbf{S}^*, \widehat{\mathbf{Z}}^{(t)}) - \nabla_\mathbf{S} q(\mathbf{S}^*, \mathbf{Z}^*)\|_F^2 \\
&= 2\|\nabla_\mathbf{S} q(\widehat{\mathbf{S}}^{(t)}, \widehat{\mathbf{Z}}^{(t)}) - \nabla_\mathbf{S} q(\mathbf{S}^*, \widehat{\mathbf{Z}}^{(t)})\|_F^2 + 2\|\nabla_\mathbf{S} p(\mathbf{S}^*, \widehat{\mathbf{L}}^{(t)}) - \nabla_\mathbf{S} p(\mathbf{S}^*, \mathbf{L}^*)\|_F^2 \\
&\leq 2\|\nabla_\mathbf{S} q(\widehat{\mathbf{S}}^{(t)}, \widehat{\mathbf{Z}}^{(t)}) - \nabla_\mathbf{S} q(\mathbf{S}^*, \widehat{\mathbf{Z}}^{(t)})\|_F^2 + 2\gamma_2^2 \|\widehat{\mathbf{L}}^{(t)} - \mathbf{L}^*\|_F^2,
\end{aligned} \tag{B.5}$$

where the first inequality is due to $(a+b)^2 \leq 2a^2 + 2b^2$, the equality is due $q(\mathbf{S}, \mathbf{Z}) = p(\mathbf{S}, \mathbf{Z}\mathbf{Z}^\top) = p(\mathbf{S}, \mathbf{L})$, and the last inequality is by the first-order stability property, i.e., Lemma B.2, where $\gamma_2 = 8\nu^2$. Submitting (B.2), (B.3), (B.4) and (B.5) into (B.1) yields

$$\begin{aligned}
\|\mathbf{S}^{(t+0.5)} - \mathbf{S}^*\|_F^2 &\leq \left(1 - \frac{2\eta\mu L}{L + \mu}\right) \|\widehat{\mathbf{S}}^{(t)} - \mathbf{S}^*\|_F^2 + 2\eta\left(\eta - \frac{1}{L + \mu}\right) \|\nabla_\mathbf{S} q(\widehat{\mathbf{S}}^{(t)}, \widehat{\mathbf{Z}}^{(t)}) - \nabla_\mathbf{S} q(\mathbf{S}^*, \widehat{\mathbf{Z}}^{(t)})\|_F^2 \\
&\quad - \frac{\eta}{2\nu^2} \|\widehat{\mathbf{L}}^{(t)} - \mathbf{L}^*\|_F \cdot \|\widehat{\mathbf{S}}^{(t)} - \mathbf{S}^*\|_F + 2\eta^2 \gamma_2^2 \|\widehat{\mathbf{L}}^{(t)} - \mathbf{L}^*\|_F^2.
\end{aligned} \tag{B.6}$$

Noting that $\|\widehat{\mathbf{L}}^{(t)} - \mathbf{L}^*\|_F \leq (R + \sqrt{\sigma_{\max}}) d(\widehat{\mathbf{Z}}^{(t)}, \mathbf{Z}^*) \leq 5/4 \sqrt{\sigma_{\max}} d(\widehat{\mathbf{Z}}^{(t)}, \mathbf{Z}^*)$, by setting $\eta \leq 1/(L + \mu)$ we have

$$\|\mathbf{S}^{(t+0.5)} - \mathbf{S}^*\|_F^2 \leq \left(1 - \frac{2\eta\mu L}{L + \mu}\right) \|\widehat{\mathbf{S}}^{(t)} - \mathbf{S}^*\|_F^2 + \frac{25\eta^2 \gamma_2^2 \sigma_{\max}}{8} d^2(\widehat{\mathbf{Z}}^{(t)}, \mathbf{Z}^*). \tag{B.7}$$

$\square$

## B.2 Proof of Lemma A.2

*Proof.* Based on the definition in (4.3) we denote

$$\bar{\mathbf{Z}}^{(t)} = \underset{\widetilde{\mathbf{Z}} \in \mathcal{U}}{\text{argmin}} \|\widehat{\mathbf{Z}}^{(t)} - \widetilde{\mathbf{Z}}\|_F,$$

which implies $d(\widehat{\mathbf{Z}}^{(t)}, \mathbf{Z}^*) = \min_{\widetilde{\mathbf{Z}} \in \mathcal{U}} \|\widehat{\mathbf{Z}}^{(t)} - \widetilde{\mathbf{Z}}\|_F = \|\widehat{\mathbf{Z}}^{(t)} - \bar{\mathbf{Z}}^{(t)}\|_F$. Thus by defining $\mathbf{Z}^{(t+1)} = \widehat{\mathbf{Z}}^{(t)} - \eta' \nabla_{\mathbf{Z}} q(\widehat{\mathbf{S}}^{(t)}, \widehat{\mathbf{Z}}^{(t)})$ as the population version of $\widehat{\mathbf{Z}}^{(t+1)}$, we have

$$d(\mathbf{Z}^{(t+1)}, \mathbf{Z}^*) = \min_{\widetilde{\mathbf{Z}} \in \mathcal{U}} \|\mathbf{Z}^{(t+1)} - \widetilde{\mathbf{Z}}\|_F \leq \|\mathbf{Z}^{(t+1)} - \bar{\mathbf{Z}}^{(t)}\|_F,$$

it follows that

$$\begin{aligned} d^2(\mathbf{Z}^{(t+1)}, \mathbf{Z}^*) &\leq \|\widehat{\mathbf{Z}}^{(t)} - \eta' \nabla_{\mathbf{Z}} q(\widehat{\mathbf{S}}^{(t)}, \widehat{\mathbf{Z}}^{(t)}) - \bar{\mathbf{Z}}^{(t)}\|_F^2 \\ &= d^2(\widehat{\mathbf{Z}}^{(t)}, \mathbf{Z}^*) - 2\eta' \underbrace{\langle \nabla_{\mathbf{Z}} q(\widehat{\mathbf{S}}^{(t)}, \widehat{\mathbf{Z}}^{(t)}), \widehat{\mathbf{Z}}^{(t)} - \bar{\mathbf{Z}}^{(t)} \rangle}_{I_1} + \eta'^2 \underbrace{\|\nabla_{\mathbf{Z}} q(\widehat{\mathbf{S}}^{(t)}, \widehat{\mathbf{Z}}^{(t)})\|_F^2}_{I_2}. \end{aligned} \quad (\text{B.8})$$

For term $I_1$ in (B.8), note that we have $\nabla_{\mathbf{Z}} q(\widehat{\mathbf{S}}^{(t)}, \widehat{\mathbf{Z}}^{(t)}) = [\nabla_{\mathbf{L}} p(\widehat{\mathbf{S}}^{(t)}, \widehat{\mathbf{L}}^{(t)})] \widehat{\mathbf{Z}}^{(t)}$, $\widehat{\mathbf{L}}^{(t)} = \widehat{\mathbf{Z}}^{(t)} [\widehat{\mathbf{Z}}^{(t)}]^\top$ and $\mathbf{L}^* = \bar{\mathbf{Z}}^{(t)} [\bar{\mathbf{Z}}^{(t)}]^\top$. It follows that

$$\begin{aligned} &\langle \nabla_{\mathbf{Z}} q(\widehat{\mathbf{S}}^{(t)}, \widehat{\mathbf{Z}}^{(t)}), \widehat{\mathbf{Z}}^{(t)} - \bar{\mathbf{Z}}^{(t)} \rangle = \langle \nabla_{\mathbf{L}} p(\widehat{\mathbf{S}}^{(t)}, \widehat{\mathbf{L}}^{(t)}), \widehat{\mathbf{Z}}^{(t)} [\widehat{\mathbf{Z}}^{(t)} - \bar{\mathbf{Z}}^{(t)}]^\top \rangle \\ &= \langle \nabla_{\mathbf{L}} p(\widehat{\mathbf{S}}^{(t)}, \mathbf{L}^*), \widehat{\mathbf{Z}}^{(t)} [\widehat{\mathbf{Z}}^{(t)} - \bar{\mathbf{Z}}^{(t)}]^\top \rangle + \langle \nabla_{\mathbf{L}} p(\widehat{\mathbf{S}}^{(t)}, \widehat{\mathbf{L}}^{(t)}) - \nabla_{\mathbf{L}} p(\widehat{\mathbf{S}}^{(t)}, \mathbf{L}^*), \widehat{\mathbf{Z}}^{(t)} [\widehat{\mathbf{Z}}^{(t)} - \bar{\mathbf{Z}}^{(t)}]^\top \rangle \\ &= \underbrace{\frac{1}{2} \langle \nabla_{\mathbf{L}} p(\widehat{\mathbf{S}}^{(t)}, \mathbf{L}^*), \widehat{\mathbf{L}}^{(t)} - \mathbf{L}^* + [\widehat{\mathbf{Z}}^{(t)} - \bar{\mathbf{Z}}^{(t)}][\widehat{\mathbf{Z}}^{(t)} - \bar{\mathbf{Z}}^{(t)}]^\top \rangle}_{I_{11}} + \underbrace{\frac{1}{2} \langle \nabla_{\mathbf{L}} p(\widehat{\mathbf{S}}^{(t)}, \widehat{\mathbf{L}}^{(t)}) - \nabla_{\mathbf{L}} p(\widehat{\mathbf{S}}^{(t)}, \mathbf{L}^*), \widehat{\mathbf{L}}^{(t)} - \mathbf{L}^* \rangle}_{I_{12}} \\ &+ \underbrace{\frac{1}{2} \langle \nabla_{\mathbf{L}} p(\widehat{\mathbf{S}}^{(t)}, \widehat{\mathbf{L}}^{(t)}) - \nabla_{\mathbf{L}} p(\widehat{\mathbf{S}}^{(t)}, \mathbf{L}^*), [\widehat{\mathbf{Z}}^{(t)} - \bar{\mathbf{Z}}^{(t)}][\widehat{\mathbf{Z}}^{(t)} - \bar{\mathbf{Z}}^{(t)}]^\top \rangle}_{I_{13}}. \end{aligned} \quad (\text{B.9})$$

We first bound term $I_{11}$ in (B.9). Noting that $\nabla_{\mathbf{L}} p(\mathbf{S}, \mathbf{L}) = \nabla_{\boldsymbol{\Omega}} p(\boldsymbol{\Omega})$, where $\boldsymbol{\Omega} = \mathbf{S} + \mathbf{L}$ and $\mathbf{L} = \mathbf{Z}\mathbf{Z}^\top$, we obtain

$$\begin{aligned} I_{11} &= \frac{1}{2} \langle \nabla_{\mathbf{L}} p(\widehat{\mathbf{S}}^{(t)}, \mathbf{L}^*), \widehat{\mathbf{L}}^{(t)} - \mathbf{L}^* + [\widehat{\mathbf{Z}}^{(t)} - \bar{\mathbf{Z}}^{(t)}][\widehat{\mathbf{Z}}^{(t)} - \bar{\mathbf{Z}}^{(t)}]^\top \rangle \\ &= \frac{1}{2} \langle \nabla_{\boldsymbol{\Omega}} p(\widehat{\mathbf{S}}^{(t)} + \mathbf{L}^*) - \nabla_{\boldsymbol{\Omega}} p(\boldsymbol{\Omega}^*), \widehat{\mathbf{L}}^{(t)} - \mathbf{L}^* + [\widehat{\mathbf{Z}}^{(t)} - \bar{\mathbf{Z}}^{(t)}][\widehat{\mathbf{Z}}^{(t)} - \bar{\mathbf{Z}}^{(t)}]^\top \rangle, \end{aligned}$$

where we used the fact that $\nabla_{\boldsymbol{\Omega}} p(\boldsymbol{\Omega}^*) = \mathbf{0}$. Applying mean value theorem yields

$$\begin{aligned} I_{11} &= 1/2 \text{vec}(\widehat{\mathbf{S}}^{(t)} - \mathbf{S}^*)^\top \nabla_{\boldsymbol{\Omega}}^2 p(\mathbf{L}^* + (1-t)\mathbf{S}^* + t\widehat{\mathbf{S}}^{(t)}) \text{vec}(\widehat{\mathbf{L}}^{(t)} - \mathbf{L}^* + [\widehat{\mathbf{Z}}^{(t)} - \bar{\mathbf{Z}}^{(t)}][\widehat{\mathbf{Z}}^{(t)} - \bar{\mathbf{Z}}^{(t)}]^\top) \\ &\geq 1/2 \lambda_{\min}(\nabla_{\boldsymbol{\Omega}}^2 p(\boldsymbol{\Omega}^* + t(\widehat{\mathbf{S}}^{(t)} - \mathbf{S}^*))) \|\widehat{\mathbf{S}}^{(t)} - \mathbf{S}^*\|_F \cdot \|\widehat{\mathbf{L}}^{(t)} - \mathbf{L}^* + [\widehat{\mathbf{Z}}^{(t)} - \bar{\mathbf{Z}}^{(t)}][\widehat{\mathbf{Z}}^{(t)} - \bar{\mathbf{Z}}^{(t)}]^\top\|_F, \end{aligned} \quad (\text{B.10})$$

for some $t \in (0,1)$. Simple calculation yields

$$\begin{aligned} \lambda_{\min}(\nabla_{\boldsymbol{\Omega}}^2 p(\boldsymbol{\Omega}^* + t(\widehat{\mathbf{S}}^{(t)} - \mathbf{S}^*))) &= \lambda_{\min}((\boldsymbol{\Omega}^* + t(\widehat{\mathbf{S}}^{(t)} - \mathbf{S}^*))^{-1} \otimes (\boldsymbol{\Omega}^* + t(\widehat{\mathbf{S}}^{(t)} - \mathbf{S}^*)^{-1}) \\ &= \|\boldsymbol{\Omega}^* + t(\widehat{\mathbf{S}}^{(t)} - \mathbf{S}^*)\|_2^{-2} \\ &\geq \frac{1}{4\nu^2}. \end{aligned} \quad (\text{B.11})$$

Thus, combining (B.10) and (B.11) we obtain

$$I_{11} \geq \frac{1}{8\nu^2} \|\widehat{\mathbf{S}}^{(t)} - \mathbf{S}^*\|_F \cdot \|\widehat{\mathbf{L}}^{(t)} - \mathbf{L}^* + [\widehat{\mathbf{Z}}^{(t)} - \bar{\mathbf{Z}}^{(t)}][\widehat{\mathbf{Z}}^{(t)} - \bar{\mathbf{Z}}^{(t)}]^\top\|_F. \quad (\text{B.12})$$

Next, since by Lemma B.1 $p(\widehat{\mathbf{S}}^{(t)}, \mathbf{L})$ is $\mu$-strongly convex and $L$-smooth with respect to $\mathbf{L}$ with $\mu = 1/(4\nu^2)$ and $L = 4\nu^2$, by Lemma D.1 we further obtain

$$I_{12} \geq \frac{\mu L}{2(\mu+L)} \|\widehat{\mathbf{L}}^{(t)} - \mathbf{L}^*\|_F^2 + \frac{1}{2(\mu+L)} \|\nabla_{\mathbf{L}} p(\widehat{\mathbf{S}}^{(t)}, \widehat{\mathbf{L}}^{(t)}) - \nabla_{\mathbf{L}} p(\widehat{\mathbf{S}}^{(t)}, \mathbf{L}^*)\|_F^2. \tag{B.13}$$

For term $I_{13}$ in (B.9), we have

$$\begin{aligned} I_{13} &\geq -\frac{1}{2}\|\nabla_{\mathbf{L}} p(\widehat{\mathbf{S}}^{(t)}, \widehat{\mathbf{L}}^{(t)}) - \nabla_{\mathbf{L}} p(\widehat{\mathbf{S}}^{(t)}, \mathbf{L}^*)\|_F \cdot \|\bar{\mathbf{Z}}^{(t)} - \widehat{\mathbf{Z}}^{(t)}\|_F^2 \\ &\geq -\frac{1}{4c}\|\nabla_{\mathbf{L}} p(\widehat{\mathbf{S}}^{(t)}, \widehat{\mathbf{L}}^{(t)}) - \nabla_{\mathbf{L}} p(\widehat{\mathbf{S}}^{(t)}, \mathbf{L}^*)\|_F^2 - \frac{c}{4}\|\bar{\mathbf{Z}}^{(t)} - \widehat{\mathbf{Z}}^{(t)}\|_F^4, \end{aligned} \tag{B.14}$$

where in the second inequality we used the inequality $2ab \leq a^2/c + cb^2$ for any $c > 0$.

Now we turn to term $I_2$ in (B.8). Recall that $\nabla_{\mathbf{Z}} q(\mathbf{S}, \mathbf{Z}) = [\nabla_{\mathbf{L}} p(\mathbf{S}, \mathbf{L})]\mathbf{Z}$. We have

$$\begin{aligned} \|\nabla_{\mathbf{Z}} q(\widehat{\mathbf{S}}^{(t)}, \widehat{\mathbf{Z}}^{(t)})\|_F^2 &\leq 2\|[\nabla_{\mathbf{L}} p(\widehat{\mathbf{S}}^{(t)}, \widehat{\mathbf{L}}^{(t)}) - \nabla_{\mathbf{L}} p(\widehat{\mathbf{S}}^{(t)}, \mathbf{L}^*)]\widehat{\mathbf{Z}}^{(t)}\|_F^2 + 2\|[\nabla_{\mathbf{L}} p(\widehat{\mathbf{S}}^{(t)}, \mathbf{L}^*) - \nabla_{\mathbf{L}} p(\mathbf{S}^*, \mathbf{L}^*)]\widehat{\mathbf{Z}}^{(t)}\|_F^2 \\ &\leq 2\|\nabla_{\mathbf{L}} p(\widehat{\mathbf{S}}^{(t)}, \widehat{\mathbf{L}}^{(t)}) - \nabla_{\mathbf{L}} p(\widehat{\mathbf{S}}^{(t)}, \mathbf{L}^*)\|_F^2 \cdot \|\widehat{\mathbf{Z}}^{(t)}\|_2^2 + 2\gamma_1^2 \|\widehat{\mathbf{S}}^{(t)} - \mathbf{S}^*\|_F^2 \cdot \|\widehat{\mathbf{Z}}^{(t)}\|_2^2 \\ &\leq \frac{25\sigma_{\max}}{8}\|\nabla_{\mathbf{L}} p(\widehat{\mathbf{S}}^{(t)}, \widehat{\mathbf{L}}^{(t)}) - \nabla_{\mathbf{L}} p(\widehat{\mathbf{S}}^{(t)}, \mathbf{L}^*)\|_F^2 + \frac{25\gamma_1^2 \sigma_{\max}}{8}\|\widehat{\mathbf{S}}^{(t)} - \mathbf{S}^*\|_F^2, \end{aligned} \tag{B.15}$$

where the second inequality is due to Lemma B.2 with $\gamma_1 = 8\nu^2$, and the last ineuqlity is due to $\|\widehat{\mathbf{Z}}^{(t)}\|_2 \leq \|\mathbf{Z}^*\|_2 + d(\widehat{\mathbf{Z}}^{(t)}, \mathbf{Z}^*) \leq R + \sqrt{\sigma_{\max}} \leq 5/4\sqrt{\sigma_{\max}}$.

Thus submitting (B.12), (B.13), (B.14) and (B.15) into (B.8) yields

$$\begin{aligned} d^2(\mathbf{Z}^{(t+1)}, \mathbf{Z}^*) &\leq \left(1 - \frac{2\eta'(\sqrt{2}-1)\sigma_{\min}\mu L}{L+\mu}\right) d^2(\widehat{\mathbf{Z}}^{(t)}, \mathbf{Z}^*) + \frac{c\eta'}{2}\|\bar{\mathbf{Z}}^{(t)} - \widehat{\mathbf{Z}}^{(t)}\|_F^4 + \frac{25\eta'^2\gamma_1^2\sigma_{\max}}{8}\|\widehat{\mathbf{S}}^{(t)} - \mathbf{S}^*\|_F^2 \\ &\quad + \left(\frac{25\eta'^2\sigma_{\max}}{8} + \frac{\eta'}{2c} - \frac{\eta'}{L+\mu}\right)\|\nabla_{\mathbf{L}} p(\widehat{\mathbf{S}}^{(t)}, \widehat{\mathbf{L}}^{(t)}) - \nabla_{\mathbf{L}} p(\widehat{\mathbf{S}}^{(t)}, \mathbf{L}^*)\|_F^2 \\ &\quad - \frac{\eta'}{4\nu^2}\|\widehat{\mathbf{S}}^{(t)} - \mathbf{S}^*\|_F \cdot \|\widehat{\mathbf{L}}^{(t)} - \mathbf{L}^* + [\widehat{\mathbf{Z}}^{(t)} - \bar{\mathbf{Z}}^{(t)}][\widehat{\mathbf{Z}}^{(t)} - \bar{\mathbf{Z}}^{(t)}]^\top\|_F \\ &\leq \left(1 - \frac{2\eta'(\sqrt{2}-1)\sigma_{\min}\mu L}{L+\mu} + \frac{\eta' R^2 (L+\mu)}{2}\right) d^2(\widehat{\mathbf{Z}}^{(t)}, \mathbf{Z}^*) + \frac{25\eta'^2\gamma_1^2\sigma_{\max}}{8}\|\widehat{\mathbf{S}}^{(t)} - \mathbf{S}^*\|_F^2, \end{aligned} \tag{B.16}$$

where in the first inequality we used the conclusion in Lemma D.4 and that $\sigma_{\min}(\mathbf{Z}^*) = \sqrt{\sigma_{\min}}$; in the second inequality we chose $c = L+\mu$ and used the condition that $\eta' \leq 4/[25(L+\mu)\sigma_{\max}]$. By our condition that $R \leq \sqrt{\sigma_{\min}}/(6.5\nu^2)$, we get $R^2 \leq 3\sigma_{\min}/(125\nu^4) \leq (4\sqrt{2}-5)\sigma_{\min}\mu L/(L+\mu)^2$, which immediately implies

$$d^2(\mathbf{Z}^{(t+1)}, \mathbf{Z}^*) \leq \left(1 - \frac{\eta'\sigma_{\min}\mu L}{2(L+\mu)}\right) d^2(\widehat{\mathbf{Z}}^{(t)}, \mathbf{Z}^*) + \frac{25\eta'^2\gamma_1^2\sigma_{\max}}{8}\|\widehat{\mathbf{S}}^{(t)} - \mathbf{S}^*\|_F^2, \tag{B.17}$$

which completes the proof. $\square$

### B.3 Proof of Lemma A.4

Now we are going to prove the lemma of statistical errors.

*Proof.* This lemma has two parts: one is the statistical error for the derivatives of loss functions with respect to $\mathbf{S}$, and the other one with respect to $\mathbf{Z}$. We first deal with $\mathbf{S}$.

**Part 1:** Taking derivative of $q(\mathbf{S}, \mathbf{Z})$ with respect to $\mathbf{S}$ while fixing $\mathbf{Z}$, we have

$$\nabla_{\mathbf{S}} q(\mathbf{S}, \mathbf{Z}) = \mathbf{\Sigma}^* - (\mathbf{S} + \mathbf{Z}\mathbf{Z}^\top)^{-1}.$$

Take derivative of $q_n(\mathbf{S}, \mathbf{Z})$ with respect to $\mathbf{S}$ while fixing $\mathbf{Z}$, we have

$$\nabla_{\mathbf{S}} q_n(\mathbf{S}, \mathbf{Z}) = \widehat{\mathbf{\Sigma}} - (\mathbf{S} + \mathbf{Z}\mathbf{Z}^\top)^{-1} = \frac{1}{n} \sum_{i=1}^n \mathbf{X}_i \mathbf{X}_i^\top - (\mathbf{S} + \mathbf{Z}\mathbf{Z}^\top)^{-1}.$$

Thus by Lemma D.2, we obtain

$$\left\| \nabla_{\mathbf{S}} q_n(\mathbf{S}, \mathbf{Z}) - \nabla_{\mathbf{S}} q(\mathbf{S}, \mathbf{Z}) \right\|_{\infty,\infty} = \left\| \frac{1}{n} \sum_{i=1}^n \mathbf{X}_i \mathbf{X}_i^\top - \mathbf{\Sigma}^* \right\|_{\infty,\infty} \leq C\nu \sqrt{\frac{\log d}{n}} \tag{B.18}$$

holds with probability at least $1 - C/d$.

**Part 2:** Taking derivative of $q(\mathbf{S}, \mathbf{Z})$ with respect to $\mathbf{Z}$ while fixing $\mathbf{S}$, we have

$$\nabla_{\mathbf{Z}} q(\mathbf{S}, \mathbf{Z}) = 2\mathbf{\Sigma}^* \mathbf{Z} - 2(\mathbf{S} + \mathbf{Z}\mathbf{Z}^\top)^{-1} \mathbf{Z}.$$

Taking derivative of $q_n(\mathbf{S}, \mathbf{Z})$ with respect to $\mathbf{Z}$ while fixing $\mathbf{S}$, we have

$$\nabla_{\mathbf{Z}} q_n(\mathbf{S}, \mathbf{Z}) = 2\widehat{\mathbf{\Sigma}} \mathbf{Z} - 2(\mathbf{S} + \mathbf{Z}\mathbf{Z}^\top)^{-1} \mathbf{Z}.$$

Then by transformation of norm, we have

$$\left\| \nabla_{\mathbf{Z}} q_n(\mathbf{S}, \mathbf{Z}) - \nabla_{\mathbf{Z}} q(\mathbf{S}, \mathbf{Z}) \right\|_F = 2 \left\| (\widehat{\mathbf{\Sigma}} - \mathbf{\Sigma}^*) \mathbf{Z} \right\|_F \leq 2 \left\| \frac{1}{n} \sum_{i=1}^n \mathbf{X}_i \mathbf{X}_i^\top - \mathbf{\Sigma}^* \right\|_2 \cdot \|\mathbf{Z}\|_F.$$

Since $\|\mathbf{Z}\|_F \leq \|\mathbf{Z} - \bar{\mathbf{Z}}\|_F + \|\bar{\mathbf{Z}}\|_F$ and $\|\mathbf{Z} - \bar{\mathbf{Z}}\|_F = d(\mathbf{Z}, \mathbf{Z}^*) \leq R$, $\|\bar{\mathbf{Z}}\|_2 = \|\mathbf{Z}^*\|_2 \leq \sqrt{\sigma_{\max}}$, we have $\|\mathbf{Z}\|_F \leq R + \sqrt{r\sigma_{\max}}$. Lemma D.3 shows that we have

$$\left\| \frac{1}{n} \sum_{i=1}^n \mathbf{X}_i \mathbf{X}_i^\top - \mathbf{\Sigma}^* \right\|_2 \leq C_0 \nu \sqrt{\frac{d}{n}}$$

with probability at least $1 - C'/d$. It immediately follows that

$$\left\| \nabla_{\mathbf{Z}} q_n(\mathbf{S}, \mathbf{Z}) - \nabla_{\mathbf{Z}} q(\mathbf{S}, \mathbf{Z}) \right\|_F \leq C' \nu \sqrt{\sigma_{\max}} \sqrt{\frac{rd}{n}}$$

holds with probability at least $1 - C'/d$. □

### B.4 Proof of Lemma A.5

*Proof.* By definition we have $\|\mathbf{A}\|_2 = \sup_{\|\mathbf{x}\|_2 = 1} \mathbf{x}^\top \mathbf{A} \mathbf{x}$. Note that

$$\mathbf{x}^\top \mathbf{A} \mathbf{x} = \langle \mathbf{x}, \mathbf{A}\mathbf{x} \rangle = \langle \mathbf{x}\mathbf{x}^\top, \mathbf{A} \rangle \leq \|\mathbf{x}\mathbf{x}^\top\|_F \cdot \|\mathbf{A}\|_F \leq \sqrt{s_0} \|\mathbf{A}\|_{\infty,\infty},$$

where in the last inequality we use the fact that $\|\mathbf{x}\mathbf{x}^\top\|_F = 1$. □

# C Proof of Additional Lemmas

## C.1 Proof of Lemma B.1

*Proof.* We first show the strong convexity and smoothness with respect to $\mathbf{S}$. Taking derivative of $p(\mathbf{S}, \mathbf{L}^*)$ with respect to $\mathbf{S}$ while fixing $\mathbf{L}^*$ and denoting the gradient as $\nabla_{\mathbf{S}} p(\mathbf{S}, \mathbf{L}^*)$, we have

$$\nabla_{\mathbf{S}} p(\mathbf{S}, \mathbf{L}^*) = \mathbf{\Sigma}^* - (\mathbf{S} + \mathbf{L}^*)^{-1}.$$

Further, taking the second order derivative with respect to $\mathbf{S}$, we get

$$\nabla_{\mathbf{S}}^2 p(\mathbf{S}, \mathbf{L}^*) = (\mathbf{S} + \mathbf{L}^*)^{-1} \otimes (\mathbf{S} + \mathbf{L}^*)^{-1}. \tag{C.1}$$

For any $\mathbf{S} \in \mathbb{B}_F(\mathbf{S}^*, R)$, we define

$$\mathcal{E}(\mathbf{S}) = \langle \nabla_{\mathbf{S}} p(\mathbf{S}, \mathbf{L}^*) - \nabla_{\mathbf{S}} p(\mathbf{S}^*, \mathbf{L}^*), \mathbf{S} - \mathbf{S}^* \rangle. \tag{C.2}$$

Applying mean value theorem to (C.2), we obtain

$$\mathcal{E}(\mathbf{S}) \geq \lambda_{\min}(\nabla_{\mathbf{S}}^2 p(\mathbf{S}^* + \theta(\mathbf{S} - \mathbf{S}^*), \mathbf{L}^*)) \|\mathbf{S} - \mathbf{S}^*\|_F^2 = \lambda_{\max}(\mathbf{S}^* + \theta(\mathbf{S} - \mathbf{S}^*) + \mathbf{L}^*)^{-2} \|\mathbf{S} - \mathbf{S}^*\|_F^2, \tag{C.3}$$

for some $\theta \in [0, 1]$, where in the last equality we use the property of Kronecker product. By triangle inequality we have

$$\lambda_{\max}(\mathbf{S}^* + \theta(\mathbf{S} - \mathbf{S}^*) + \mathbf{L}^*) \leq \|\mathbf{S}^* + \mathbf{L}^*\|_2 + \theta \|\mathbf{S} - \mathbf{S}^*\|_2 \leq \nu + R \leq 2\nu, \tag{C.4}$$

where the last inequality is because we have $R \leq 1/\nu \leq \nu$ by definition. Combining (C.3) and (C.4) yields

$$\mathcal{E}(\mathbf{S}) \geq \frac{1}{4\nu^2} \|\mathbf{S} - \mathbf{S}^*\|_F^2,$$

which immediately implies that $q(\mathbf{S}, \mathbf{L}^*)$ is $\mu$-strongly convex with respect to $\mathbf{S}$, where $\mu = 1/(4\nu^2)$.

Note that for $\mathcal{E}(\mathbf{S})$ defined in (C.2), we also have

$$\mathcal{E}(\mathbf{S}) \leq \lambda_{\max}(\nabla_{\mathbf{S}}^2 p(\mathbf{S}^* + \theta(\mathbf{S} - \mathbf{S}^*), \mathbf{L}^*)) \|\mathbf{S} - \mathbf{S}^*\|_F^2 = \lambda_{\min}(\mathbf{S}^* + \theta(\mathbf{S} - \mathbf{S}^*) + \mathbf{L}^*)^{-2} \|\mathbf{S} - \mathbf{S}^*\|_F^2, \tag{C.5}$$

For any $\mathbf{x} \in \mathbb{R}^d$ such that $\|\mathbf{x}\|_2 = 1$, we have

$\mathbf{x}^\top (\mathbf{S}^* + \theta(\mathbf{S} - \mathbf{S}^*) + \mathbf{L}^*) \mathbf{x} = (1-\theta) \mathbf{x}^\top (\mathbf{S} - \mathbf{S}^*) \mathbf{x} + \mathbf{x}(\mathbf{S}^* + \mathbf{L}^*)\mathbf{x} \geq -(1-\theta)|\mathbf{x}^\top (\mathbf{S} - \mathbf{S}^*) \mathbf{x}| + \mathbf{x}(\mathbf{S}^* + \mathbf{L}^*)\mathbf{x},$

where the last inequality is due to $0 \leq \theta \leq 1$. Taking minimization over $\mathbf{x}$ on both side of the inequality above, we have

$$\lambda_{\min}(\mathbf{S}^* + \theta(\mathbf{S} - \mathbf{S}^*) + \mathbf{L}^*) = \min_{\|\mathbf{x}\|_2 = 1} \mathbf{x}^\top (\mathbf{S}^* + \theta(\mathbf{S} - \mathbf{S}^*) + \mathbf{L}^*) \mathbf{x}$$

$$\geq (1-\theta) \min_{\|\mathbf{x}\|_2=1} \left\{ -|\mathbf{x}^\top (\mathbf{S} - \mathbf{S}^*) \mathbf{x}| \right\} + \min_{\|\mathbf{x}\|_2=1} \mathbf{x}(\mathbf{S}^* + \mathbf{L}^*)\mathbf{x}$$

$$\geq \frac{1}{\nu} - R \geq \frac{1}{2\nu},$$

where in the last inequality we use the fact $\mathbf{S} \in \mathbb{B}_F(\mathbf{S}^*, R)$ and $R \leq 1/(2\nu)$. Then it follows that

$$\mathcal{E}(\mathbf{S}) \leq 4\nu^2 \|\mathbf{S}' - \mathbf{S}\|_F^2,$$

which immediately implies that $p(\mathbf{S}, \mathbf{L}^*)$ is $L$-smooth with respect to $\mathbf{S}$, and $L = 4\nu^2$.

Since $p(\mathbf{S}, \mathbf{L})$ is symmetric in $\mathbf{S}$ and $\mathbf{L}$, by similar proof for $\mathbf{L}$, we can show that $p(\mathbf{S}^*, \mathbf{L})$ is $\mu$ strongly-convex and $L$-smooth with respect to $\mathbf{L}$ too. $\square$

## C.2 Proof of Lemma B.2

In this subsection, we prove the first-order stability lemmas.

*Proof.* Take derivative of $p(\mathbf{S}, \mathbf{L})$ with respect to $\mathbf{S}$ while fixing $\mathbf{L}$, we have

$$\nabla_1 p(\mathbf{S}, \mathbf{L}) = \mathbf{\Sigma}^* - (\mathbf{S} + \mathbf{L})^{-1}.$$

Therefore, we have

$$\|\nabla_1 p(\mathbf{S}, \mathbf{L}) - \nabla_1 p(\mathbf{S}, \mathbf{L}^*)\|_F \leq \left\|(\mathbf{S} + \mathbf{L})^{-1} - (\mathbf{S} + \mathbf{L}^*)^{-1}\right\|_F. \tag{C.6}$$

We define $\mathbf{\Theta}^* = \mathbf{S} + \mathbf{L}^*, \mathbf{\Theta} = \mathbf{S} + \mathbf{L}$ and $\mathbf{\Delta} = \mathbf{\Theta}^* - \mathbf{\Theta} = \mathbf{L}^* - \mathbf{L}$. Then we have

$$\left\|(\mathbf{S} + \mathbf{L})^{-1} - (\mathbf{S} + \mathbf{L}^*)^{-1}\right\|_F = \left\|(\mathbf{\Theta}^* - \mathbf{\Delta})^{-1} - \mathbf{\Theta}^{*-1}\right\|_F.$$

Since $\|\mathbf{\Theta}^{*-1}\mathbf{\Delta}\|_F \leq 1$, we have the convergent matrix expansion

$$(\mathbf{\Theta}^* - \mathbf{\Delta})^{-1} = \left[\mathbf{\Theta}^*(\mathbf{I} - \mathbf{\Theta}^{*-1}\mathbf{\Delta})\right]^{-1} = \sum_{k=0}^{\infty}(\mathbf{\Theta}^{*-1}\mathbf{\Delta})^k \mathbf{\Theta}^{*-1}.$$

Define $\mathbf{J} = \sum_{k=0}^{\infty}(\mathbf{\Theta}^{*-1}\mathbf{\Delta})^k$, we have

$$\|(\mathbf{\Theta}^* - \mathbf{\Delta})^{-1} - \mathbf{\Theta}^{*-1}\|_F = \left\|\sum_{k=1}^{\infty}(\mathbf{\Theta}^{*-1}\mathbf{\Delta})^k \mathbf{\Theta}^{*-1}\right\|_F = \left\|(\mathbf{\Theta}^{*-1}\mathbf{\Delta})\mathbf{J}\mathbf{\Theta}^{*-1}\right\|_F \leq \|\mathbf{\Theta}^{*-1}\|_2^2 \cdot \|\mathbf{\Delta}\|_F \cdot \|\mathbf{J}\|_2, \tag{C.7}$$

where we use the properties of matrix norm that $\|\mathbf{A}\mathbf{B}\|_F \leq \|\mathbf{A}\|_2 \cdot \|\mathbf{B}\|_F$ and $\|\mathbf{A}\mathbf{B}\|_2 \leq \|\mathbf{A}\|_2 \cdot \|\mathbf{B}\|_2$. By sub-multiplicativity of matrix norm, we have

$$\|\mathbf{J}\|_2 \leq \sum_{k=0}^{\infty} \|\mathbf{\Theta}^{*-1}\mathbf{\Delta}\|_2^k \leq \frac{1}{1 - \|\mathbf{\Theta}^{*-1}\mathbf{\Delta}\|_2} \leq \frac{1}{1 - \|\mathbf{\Theta}^{*-1}\|_2\|\mathbf{\Delta}\|_2} \leq 2. \tag{C.8}$$

Note that we have $\|\mathbf{\Theta}^{*-1}\|_2 = \lambda_{\max}(\mathbf{\Theta}^{*-1}) = (\lambda_{\min}(\mathbf{\Theta}^*))^{-1}$. For any $\mathbf{x} \in \mathbb{R}^d$, we have

$$\begin{aligned}
\lambda_{\min}(\mathbf{\Theta}^*) &= \min_{\|\mathbf{x}\|_2 = 1} \mathbf{x}^\top (\mathbf{S} + \mathbf{L}^*) \mathbf{x} \\
&\geq \min_{\|\mathbf{x}\|_2 = 1} \left\{-\left|\mathbf{x}^\top(\mathbf{S} - \mathbf{S}^*)\mathbf{x}\right| + \mathbf{x}^\top(\mathbf{S}^* + \mathbf{L}^*)\mathbf{x}\right\} \\
&\geq -\|\mathbf{S} - \mathbf{S}^*\|_2 + \lambda_{\min}(\mathbf{\Omega}^*) \\
&\geq 1/\nu - R > \frac{1}{2\nu},
\end{aligned} \tag{C.9}$$

where we use the fact that $\|\mathbf{S} - \mathbf{S}^*\|_2 \leq \|\mathbf{S} - \mathbf{S}^*\|_F \leq R$, $\lambda_{\min}(\mathbf{\Omega}_1^*) \geq 1/\nu$ by Assumption 4.1 and $R \leq 1/(2\nu)$. Combining (C.7), (C.8) and (C.9), we have

$$\|(\mathbf{\Theta} + \mathbf{\Delta})^{-1} - \mathbf{\Theta}^{-1}\|_F \leq (2\nu)^2 \cdot \mathbf{\Delta} \cdot 2 \leq 8\nu^2 \|\mathbf{L} - \mathbf{L}^*\|_F, \tag{C.10}$$

which ends the proof. The proof for first-order stability of $\nabla_{\mathbf{L}} p(\mathbf{S}, \mathbf{L})$ is similar and omitted here. □

# D Auxiliary Lemmas

**Lemma D.1.** (Nesterov, 2013) Let $f$ be $\mu$-strongly convex and $L$-smooth. Then for any $\mathbf{x}, \mathbf{y} \in \mathbf{dom} f$, we have

$$\langle \nabla f(\mathbf{x}) - \nabla f(\mathbf{y}), \mathbf{x} - \mathbf{y} \rangle \geq \frac{\mu L}{L + \mu} \|\mathbf{x} - \mathbf{y}\|_2^2 + \frac{1}{\mu + L} \|\nabla f(\mathbf{x}) - \nabla f(\mathbf{y})\|_2^2.$$

**Lemma D.2.** (Ravikumar et al., 2011) Suppose that $\boldsymbol{X}_1, \ldots, \boldsymbol{X}_n \in \mathbb{R}^d$ are i.i.d. sub-Gaussian random vectors. Let $\boldsymbol{\Sigma}^* = \mathbb{E}[1/n \sum_{i=1}^n \boldsymbol{X}_i \boldsymbol{X}_i^\top]$, and we have that

$$\left\| \frac{1}{n} \sum_{i=1}^n \boldsymbol{X}_i \boldsymbol{X}_i^\top - \boldsymbol{\Sigma}^* \right\|_{\infty,\infty} \leq C \max_i \boldsymbol{\Sigma}_{ii}^* \sqrt{\frac{\log d}{n}}$$

holds with probability at least $1 - C/d$, where $C > 0$ is a constant.

**Lemma D.3.** (Vershynin, 2010) Suppose that $\boldsymbol{X}_1, \ldots, \boldsymbol{X}_n \in \mathbb{R}^d$ are i.i.d. sub-Gaussian random vectors. Let $\boldsymbol{\Sigma}^* = \mathbb{E}[1/n \sum_{i=1}^n \boldsymbol{X}_i \boldsymbol{X}_i^\top]$, and we have that

$$\left\| \frac{1}{n} \sum_{i=1}^n \boldsymbol{X}_i \boldsymbol{X}_i^\top - \boldsymbol{\Sigma}^* \right\|_2 \leq C \lambda_{\max}(\boldsymbol{\Sigma}^*) \sqrt{\frac{d}{n}}$$

holds with probability at least $1 - C/d$, where $C > 0$ is a constant.

**Lemma D.4.** (Tu et al., 2015) For any $\mathbf{Z}, \mathbf{Z}^* \in \mathbb{R}^{d \times r}$, we have

$$d(\mathbf{Z}, \mathbf{Z}^*) \leq \frac{1}{\sqrt{2(\sqrt{2} - 1)\sigma_{\min}(\mathbf{Z}^*)}} \left\| \mathbf{Z}\mathbf{Z}^\top - \mathbf{Z}^*\mathbf{Z}^{*\top} \right\|_F,$$

where $\sigma_{\min}(\mathbf{Z}^*)$ is the minimal nonzero singular value of $\mathbf{Z}^*$.